%% file: main.tex
\documentclass{article} % For LaTeX2e
\usepackage{collas2022_conference,times}

\input{header}
\input{symbols}

\title{Improved Policy Optimization for \\ Online Imitation Learning}

\author{J. Wilder Lavington \\
University of British Columbia \\ \texttt{jola2372@cs.ubc.ca} \\
\And % Use And to have authors side by side
Sharan Vaswani \\
Simon Fraser University \\
\texttt{vaswani.sharan@gmail.com} \\
\And % Use AND to have authors block one under the other
Mark Schmidt \\
University of British Columbia \\ \texttt{schmidtm@cs.ubc.ca} \\
}

\pgfplotsset{compat=1.17}
\collasfinalcopy % Uncomment for camera-ready version, but NOT for submission.

\begin{document}

\maketitle

\input{abstract}
\input{introduction}
\input{background}

\input{ftl}

\input{ftrl}

\input{experiments}
\input{related-work}

\input{discussion}

\input{acknowledgements}
\newpage 
\bibliography{main}
\bibliographystyle{collas2022_conference}
\newpage 
\appendix
\input{app-setup}

\input{app-proofs-ftrl}
\input{app-proofs-ftl}
\input{app-experimental-details} 

\end{document}

%% file: header.tex
\usepackage{lmodern}
\usepackage[english]{babel}
\usepackage{latexsym}
\usepackage{amsmath}
\usepackage{mathrsfs}
\usepackage{amssymb}
\usepackage{mathtools}
\usepackage{bm}
\usepackage{datetime}

\usepackage{accents}
\usepackage{tikz}
\usepackage{listings}
\usepackage{mdframed}
\usepackage{pgfplots}
\usepackage{pgfplotstable}
\usepackage{amsthm}

\usepackage{thmtools, thm-restate}
\usepackage{algorithm}
\usepackage{algpseudocode}
\usepackage{dsfont}
\usepackage{color}
\usepackage{colortbl}
\usepackage{pifont}
\usepackage{caption}
\usepackage{microtype} % improved spacing between words for easier reading
\usepackage{float}
\usepackage{xfrac} % sfrac
\usepackage{xspace}
\usepackage{booktabs}
\usepackage{blkarray}
\usepackage{graphicx}
\usepackage{subcaption}

\usepackage{hyperref}
\hypersetup{
    unicode=false,          % non-Latin characters in AcrobatÕs bookmarks
    pdftoolbar=true,        % show AcrobatÕs toolbar?
    pdfmenubar=true,        % show AcrobatÕs menu?
    pdffitwindow=false,     % window fit to page when opened
    pdfstartview={FitH},    % fits the width of the page to the window
    pdftitle={XXXXX},    % title
    pdfauthor={XXX},     % author
    pdfsubject={Bandits, Reinforcement Learning},   % subject of the document
    pdfcreator={Creator},   % creator of the document
    pdfproducer={Producer}, % producer of the document
    pdfkeywords={bandits} {reinforcement learning} {policy gradient}, % list of keywords
    pdfnewwindow=true,      % links in new window
    colorlinks=true,       % false: boxed links; true: colored links
    linkcolor=blue,          % color of internal links (change box color with linkbordercolor)
    citecolor=blue,        % color of links to bibliography
    filecolor=magenta,      % color of file links
    urlcolor=cyan           % color of external links
}
\usepackage{times}
\usepackage{nicefrac}
\usepackage{wrapfig}
\usepackage{pgfplots}
\usepackage[capitalize]{cleveref}

\theoremstyle{plain}
\newtheorem{theorem}{Theorem}[section]

\newtheorem{lemma}[theorem]{Lemma}

\theoremstyle{definition}
\newtheorem{definition}[theorem]{Definition}

\theoremstyle{remark}

\mdfdefinestyle{mdframedthmbox}{%
	leftmargin=.0\textwidth,
	rightmargin=.0\textwidth,%
	innertopmargin=0.75em,
	innerleftmargin=.5em,
	innerrightmargin=.5em,
}

\newenvironment{thmbox}
	{%
		\begin{mdframed}[style=mdframedthmbox]%
	}{% 
		\end{mdframed}%
	}

\usepackage{float}
\usepackage{tikz}
\usetikzlibrary{shapes.geometric}
\usetikzlibrary{calc}
\usetikzlibrary{shapes,arrows}
\usetikzlibrary{matrix,positioning}
\usetikzlibrary{bayesnet}
\usepackage{wrapfig}

%% file: symbols.tex
\newcommand{\cA}{\mathcal{A}}

\newcommand{\cE}{\mathcal{E}}

\newcommand{\cP}{\mathcal{P}}

\newcommand{\cS}{\mathcal{S}}

\newcommand{\cW}{\mathcal{W}}

\def\eqref#1{equation~\ref{#1}}
\def\1{\bm{1}}

\DeclareMathAlphabet{\mathsfit}{\encodingdefault}{\sfdefault}{m}{sl}
\SetMathAlphabet{\mathsfit}{bold}{\encodingdefault}{\sfdefault}{bx}{n}

\newcommand{\E}{\mathbb{E}}

\newcommand{\R}{\mathbb{R}}

\DeclareMathOperator*{\argmin}{arg\,min}

\newcommand{\etat}{\eta_{t}}
\newcommand{\xkk}{w_{t+1}}
\newcommand{\xk}{w_{t}}
\newcommand{\xs}{w_{s}}

\newcommand{\x}{w}

\newcommand{\xt}{w_{t}}
\newcommand{\xtt}{w_{t+1}}
\newcommand{\xopt}{w^{*}}

\newcommand{\lt}{l_t}

\newcommand{\li}{l_i}

\newcommand{\psit}{\psi_t}
\newcommand{\psitt}{\psi_{t+1}}

\newcommand{\gradt}[1]{\nabla l_{t}(#1)}

\newcommand{\gradi}[1]{\nabla l_{i}(#1)}

\newcommand{\Ft}{F_t}
\newcommand{\Ftt}{F_{t+1}}

\newcommand{\lambdat}{\lambda_t}
\newcommand{\lambdatt}{\lambda_{t+1}}

\newcommand{\pie}{\pi_e}

\newcommand{\pitt}{\pi_{t+1}}
\newcommand{\pit}{\pi_{t}}

\newcommand{\sigmas}{\sigma_{s}}
\newcommand{\sigmat}{\sigma_{t}}
\newcommand{\sigmak}{\sigma_{t}}

\newcommand{\norm}[1]{\left\|#1\right\|}
\newcommand{\normsq}[1]{\left\|#1\right\|^{2}}
\newcommand{\grads}[1]{\nabla l_{s}(#1)}

\newcommand{\inner}[2]{\langle #1, #2 \rangle}

\newcommand{\aligns}[1]{\begin{align*} #1 \end{align*}}

%% file: abstract.tex
\begin{abstract}
We consider online imitation learning (OIL), where the task is to find a policy that imitates the behavior of an expert via active interaction with the environment. We aim to bridge the gap between the theory and practice of policy optimization algorithms for OIL by analyzing one of the most popular OIL algorithms, DAGGER. Specifically, if the class of policies is sufficiently expressive to contain the expert policy, we prove that DAGGER achieves constant regret. Unlike previous bounds that require the losses to be strongly-convex, our result only requires the weaker assumption that the losses be strongly-convex with respect to the policy's sufficient statistics (not its parameterization). In order to ensure convergence for a wider class of policies and losses, we augment DAGGER with an additional regularization term. In particular, we propose a variant of Follow-the-Regularized-Leader (FTRL) and its adaptive variant for OIL and develop a memory-efficient implementation, which matches the memory requirements of FTL. Assuming that the loss functions are smooth and convex with respect to the parameters of the policy, we also prove that FTRL achieves constant regret for any sufficiently expressive policy class, while retaining $O(\sqrt{T})$ regret in the worst-case. We demonstrate the effectiveness of these algorithms with experiments on synthetic and high-dimensional control tasks. 
\end{abstract}

%% file: introduction.tex
\section{Introduction}
\label{sec:introduction}
Learning to make control decisions online in a stable and efficient manner is important in computer animation~\citep{motionvaes2020, 2018-MIG-autoComplete}, resource management~\citep{zhou2011optimal,IGNACIUK20101982}, robotics~\citep{andrychowicz2020learning,xie2018feedback,schaal2010learning}, and autonomous vehicles~\citep{chen2020learning,sadigh2016planning}. Online decision making has a variety of challenges: from partial-observability and asymmetric information~\citep{pmlr-v139-warrington21a,choudhury2018data}, to function approximation and bootstrapping error~\citep{van2018deep}. One common method to avoid some of these problems is through online imitation learning~\citep{ross2011reduction} (OIL). The OIL setting assumes access to an expert which is known to achieve the desired control objective (e.g. drive safely), and the task is to learn a policy that imitates the behavior of this expert through direct interaction with the environment by the learned policy. 

Although there has been substantial progress in practical algorithms for IL such as imitation learning from observations alone~\citep{kidambi2021mobile, peng2018deepmimic}, adversarial IL (AIL)~\citep{ghasemipour2020divergence, creswell2018generative, fu2018learning}, learning from imperfect experts~\citep{sun2018truncated, pmlr-v78-laskey17a,sun2017deeply} or demonstrations~\citep{rengarajan2022reinforcement, reddy2019sqil, 8463162}, and learning amortized proposals for planning~\citep{https://doi.org/10.48550/arxiv.2205.15460,fickinger2021scalable,piche2018probabilistic}, there has been relatively little work on direct policy optimization. Even in areas which sometimes provide guarantees like apprenticeship learning~\citep{shani2021online, syed2007game, abbeel2004apprenticeship}, and behavioral cloning (BC)~\citep{florence2021implicit}, there is still a large gap between theory and practice. 

For OIL, one of the most popular policy optimization algorithms, DAGGER~\citep{ross2011reduction}, minimizes the discrepancy between the learned policy and the expert over all states observed through interaction with the environment. \citet{ross2011reduction} frame the OIL problem as online convex optimization (OCO)~\citep{hazan2021introduction}, where the sequence of functions measure the discrepancy between the current policy and the expert. \citet{ross2011reduction} show that DAGGER is in fact an instance of the follow-the-leader (FTL) algorithm~\citep{hazan2007logarithmic} and inherits the FTL guarantees when the discrepancy function is strongly-convex in the parameters of the policy. The advantage of the OCO framework is that it actively models adversarial sequences of functions and ensures the resulting algorithms guard against worst case behavior. 

However, in OIL the functions are not adversarial. Instead, they are generated by a behavioral policy used to interact with the environment and the function used to measure the discrepancy between the learned and expert policies. Consequently, some OIL algorithms, including DAGGER, have good empirical performance for a broader range of function classes than suggested by the theory. Recent work by \citet{DBLP:journals/corr/abs-2007-02520} suggests that this theory-practice inconsistency stems from the use of highly expressive policy classes. Assuming that the policy class contains the expert policy,~\citet{DBLP:journals/corr/abs-2007-02520} prove that common OCO algorithms including follow-the-regularized leader~\citep{abernethy2009competing}, online gradient descent~\citep{zinkevich2003online}, and AdaGrad~\citep{duchi2011adaptive} have better worst-case performance than suggested by the existing theory. However, they (i) only focus on convex functions where modern OIL involves minimizing non-convex loss functions, and (ii) only consider the linearized variants of FTRL such as AdaGrad which are less sample-efficient then their unlinearized counter-parts because they do not take advantage of previous examples. In this work, we address these issues and make the following contributions. 

%  this gap, and provide scalable implementations of these algorithms.
% than suggested by the theory. better in practice than 
% Instead they are as a result, algorithms such as DAGGER work better in practice than suggested by theory. 
% expert and  simply aggregates all previous observations obtained from environment interactions, and % the much simpler setting of OIL
% overcoming the exploration-exploitation trade-off through 
% Crucially, this trend will continue to become more important in the future, as technologies which run city infrastructure, vehicles, and personal devices become more complex.% In all such examples, we wish to achieve some control objective, like driving safely. From a theoretical perspective,~\citet{ross2011reduction} This view reduces the OIL problem to minimizing the loss over the observed sequence of functions. and 

\subsection{Contributions}
\label{sec:contributions}
\textbf{Follow-the-Leader}: Instead of focusing on the worst-case performance of FTL for arbitrary convex functions, in~\cref{sec:ftl}, we analyze the theoretical performance of FTL (and thus DAGGER) by exploiting the specific structure in OIL problems. In particular, assuming the class of policies (which we are optimizing over) is sufficiently expressive such that it contains the expert policy, we prove FTL can achieve \emph{constant regret} for OIL problems (\cref{thm:ftl}). Unlike previous work~\citep{DBLP:journals/corr/abs-2007-02520}, this result justifies the superior empirical performance of FTL, and does not require convexity or smoothness with respect to the policy parameterization. Furthermore, we show the use of expressive policy classes can also improve the computational complexity of FTL, making it more robust to hyperparameter tuning. Our analysis shows that much of the empirical success of DAGGER might be due both to its use of specific loss functions as well as expressive policy classes. 

\begin{figure}[t] 
         \centering 
         \raisebox{-0.08\height}{
         \includegraphics[width=0.45\textwidth]{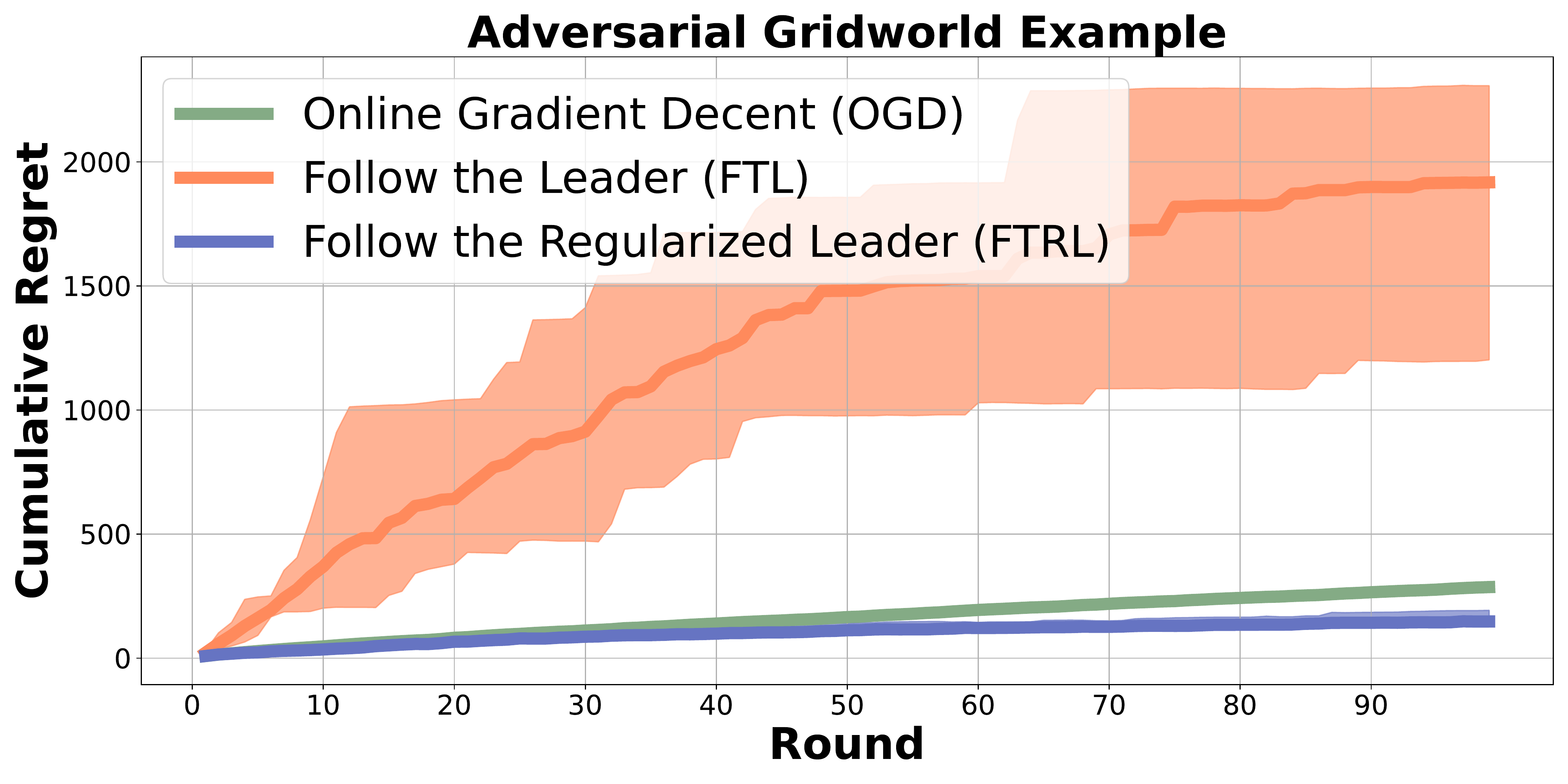}}
         \includegraphics[width=0.45\textwidth]{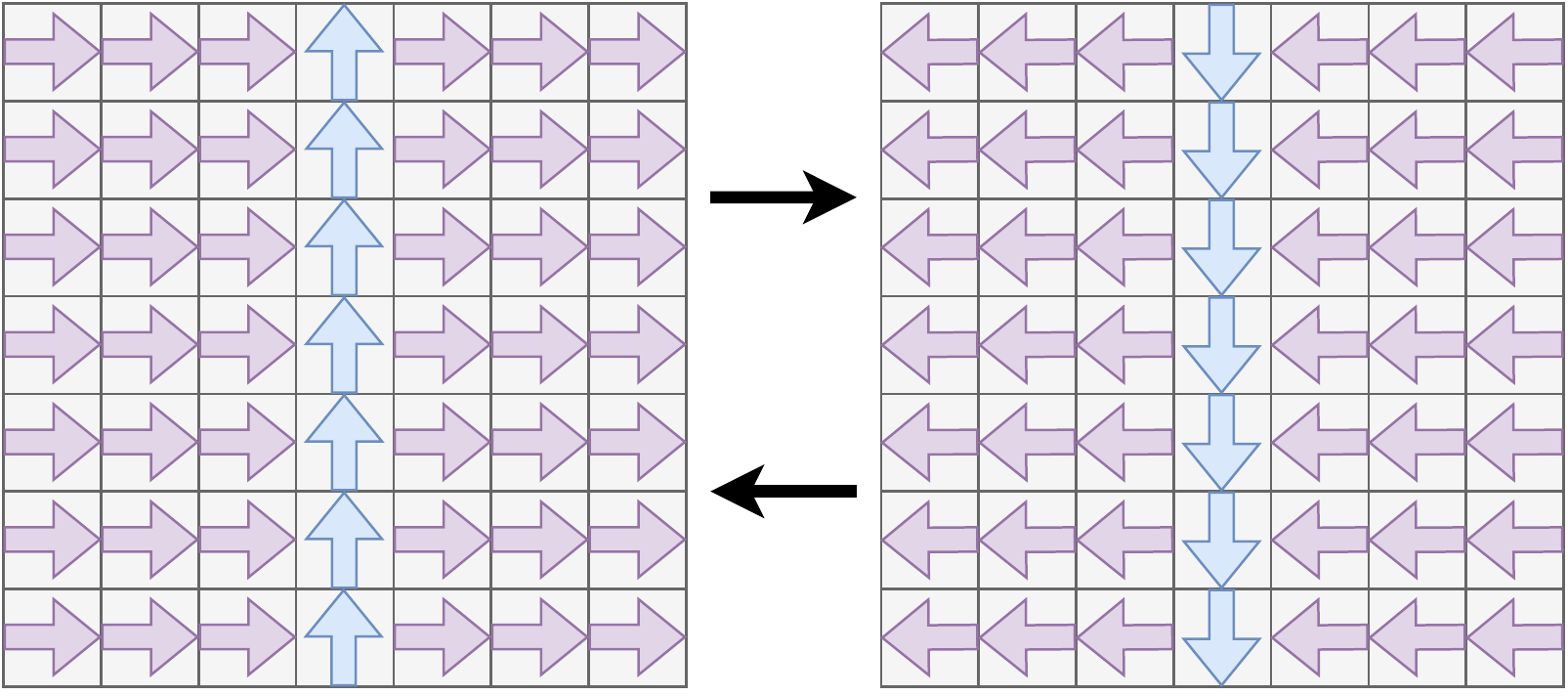}
         \caption{Adversarial Gridworld with a finite-horizon MDP and linear policy parameterization. The agent attempts to match noisy expert feedback in each round: the blue and purple arrows denoting the expert action and the two grids illustrate expert actions in alternate rounds. We observe that FTL has substantially larger cumulative regret, while OGD and FTRL demonstrate better empirical performance (details in~\cref{app:examples:gridworld}).}
    \label{fig:adversarial_grid_plots}
\end{figure}

However, when the policy class is not rich enough or the agent is not provided with sufficient state information, DAGGER can result in linear regret and poor empirical performance~\citep{pmlr-v139-warrington21a, choudhury2018data}. As a simple example, the grid-world in~\cref{fig:adversarial_grid_plots} shows how DAGGER (denoted by its corresponding online optimization algorithm FTL) might exhibit poor oscillatory behavior. This is reminiscent of the counter-example for FTL in the OCO setting in the absence of strong-convexity~\citep{shalev2012online}. In the OCO literature, adding a regularization term is the standard way to remedy such oscillatory behavior~\citep{abernethy2009competing}. 

\textbf{Follow-the-Regularized-Leader}: Analogously, we use the follow-the-regularized-leader (FTRL) algorithm~\citep{abernethy2009competing} in the OIL setting (\cref{sec:ftrl}). FTRL generalizes FTL, helping it guard against adversarial examples similar to~\cref{fig:adversarial_grid_plots}. Unlike FTL, our FTRL analysis assumes loss functions are smooth and convex with respect to the policy parameterization. In~\cref{thm:adaftrl-regret}, we prove that FTRL can obtain constant regret if the policy class is sufficiently expressive and contains the expert policy. In the absence of an expressive policy class, FTRL still results in sublinear regret, improving over FTL in this setting. However, unlike FTL which is parameter-free, FTRL requires that its regularization strength be set according to unknown problem-dependent constants, which can result in poor empirical performance. Consequently, we make use of the adaptive FTRL (AdaFTRL) algorithm~\citep{mcmahan2017survey} in the OIL setting. Using a proof technique similar to~\citet{vaswani2020adaptive,DBLP:journals/corr/abs-2007-02520, levy2018online,pmlr-v108-xie20a}, for smooth, convex loss functions, we prove that AdaFTRL obtains the same regret guarantees as FTRL. We additionally show that unlike FTRL, AdaFTRL does not require the knowledge of problem-dependent constants (\cref{thm:adaftrl-regret}). 

\textbf{Experimental Evaluation}: In~\cref{sec:experiments}, we evaluate the algorithms for both continuous~\citep{todorov2012mujoco} and discrete control~\citep{mnih2013playing}. Our experiments demonstrate the superior empirical performance of FTL and (Ada-)FTRL, methods that update the policy by utilizing all the past data (so-called ``offline'' updates). Our experiments also indicate the benefit of functional regularization, with the FTRL variants often outperforming DAGGER (FTL) in terms of either average cumulative loss or average return. An accompanying codebase can be found here: \href{https://github.com/WilderLavington/Improved-Policy-Optimization-for-Online-Imitation-Learning.git}{Improved-Policy-Optimization-for-Online-Imitation-Learning.git}. This code includes all algorithms and baselines discussed in the paper, as well as the additional experiments discussed in \cref{app:exp-details}.  

%% file: background.tex
\section{Problem Formulation}
\label{sec:mdp_background}
We consider an infinite-horizon discounted  Markov decision process (MDP)~\citep{bertsekas2019reinforcement,sutton2018reinforcement,Puterman1994} denoted by $M$, and defined by the tuple $\langle \cS, \cA, \cP, r, \rho, \gamma \rangle$ where $\cS$ is the set of states, $\cA$ is the action set, $\cP : \cS \times \cA \rightarrow \Delta_\cS$ is the transition distribution, $\rho \in \Delta_{\cS}$ is the initial state distribution and $\gamma \in [0, 1)$ is the discount factor. Here, $\Delta_\cS$ and $\Delta_\cA$ refer to the $|\cS|$-dimensional and $|\cA|$-dimensional probability densities  respectively. In reinforcement learning (RL), we wish to maximize a reward function denoted by $r : \cS \times \cA \rightarrow \mathbb{R}$. The expected discounted return or \emph{value function} of a policy $\pi : \cS \rightarrow \Delta_{\cA}$ is defined as $V^\pi(s) = \mathbb{E}_{a_0,s_1,a_1 \ldots} \left[\sum_{\tau=0}^\infty \gamma^\tau r(s_\tau, a_\tau) | s \right]$, where $a_\tau \sim \pi( \cdot| s_\tau),$ and $s_{\tau+1} \sim \cP(s_{\tau+1} | s_\tau, a_\tau)$ and $V^\pi(\rho) := \E_{s_0 \sim \rho} V^\pi(s)$. A policy $\pi$ induces a measure $d^\pi$ over states such that $d^\pi(s) = (1-\gamma) \sum_{\tau = 0}^\infty \gamma^\tau \mathrm{Pr}^{\pi}[s_{\tau} = s \mid s_0 \sim \rho]$, where $\mathrm{Pr}^{\pi}(s_{\tau} = s \mid s_0 \sim \rho)$ is the visiting probability of $s$ when playing policy $\pi$ starting from $s_0\sim \rho$. Given a class of policies $\Pi$, the objective is to return a policy that maximizes the value function, $\max_{\pi \in \Pi} V^\pi(\rho)$. 
 
\subsection{Reduction to Online Convex Optimization}
\label{sec:il-oco}
Choosing a reward function which is easy to learn from, and that achieves a desired engineering goal, can be difficult~\citep{dulac2019challenges,NIPS2017_32fdab65, sadigh2017active}. Therefore, control engineers often use expert supervision to directly learn an optimal policy. Such experts can make execution of controllers online computationally cheaper at test time~\cite{https://doi.org/10.48550/arxiv.2205.15460}, or to ``warm-start" learning for a complicated control task~\citep{liu2021improved}. In all cases, given access to such an expert policy $\pie$, the aim of OIL is to output a policy that imitates the expert. In particular, if the divergence $D : \Delta_{\cA} \times \Delta_{\cA} \rightarrow \R$ measures the discrepancy between two policy distributions (for example the KL or Wasserstein divergence), 
\begin{equation} \label{eq:objective} 
\pi^* = \min_{\pi \in \Pi} \mathop{\mathbb{E}}_{s \sim d^\pi} [D \left(\pi(\cdot|s), \pie(\cdot|s) \right) ].  
\end{equation}
That is to say, we search for a policy which minimizes the divergence between expert and agent policies. By imitating the expert, OIL intends to learn a policy that achieves a high return such that $V^{\pi^*}(\rho) \approx V^{\pi_e}(\rho)$. Since we cannot compute $d^\pi$ or differentiate through it in general, OIL iteratively samples states from $d^\pi$, and solves the following optimization problem at iteration $t \in [T]$~\citep{ross2011reduction,DBLP:journals/corr/abs-2007-02520},
\begin{equation}
\pitt = \argmin_{\pi \in \Pi} l_t(\pi) = \argmin_{\pi \in \Pi} \mathop{\mathbb{E}}_{s \sim d^{\pit}} [ D\left( \pi(\cdot|s), \pie(\cdot|s) \right) ]. \label{eq:pi-objective}
\end{equation}  %
% In an idealized setting, for a sufficiently large class of policies $\Pi$ such that $\pie \in \Pi$, and under sufficient state-space coverage for the policies $\{\pi_0, \ldots \pi_T\}$, as $T \rightarrow \infty$, $V^{\pi_T} \rightarrow V^{\pie}$. 
\!\!\!\! In this paper, we assume that $\Pi$ consists of policies that are realizable through a set of sufficient statistics, by a model parameterized with $\x \in \cW$.  We use $\pi_{\x}$ to refer to the parametric realization of $\pi$, with the choice of the policy parameterization implicit in the $\pi_{\x}$ notation. For example, a \emph{linear policy parameterization} often assumes access to features $\phi_{s,a} \in \R^d$, and assumes that there exists a $\x \in \R^d$ such that, $\pi(a|s) \propto \exp(\langle w, \phi_{s,a} \rangle)$. 
% For a given policy parameterization and MDP,~\cref{eq:pi-objective} can be rewritten as a minimization problem over a set constrained by policies which are realizable under the chosen parameterization:
% \begin{align}
% w_{t+1} & \in \argmin_{\pi_w \in \Pi_w} \mathop{\mathbb{E}}_{s \sim d^{\pi_{_{t}}}} [ D\left( \pi(\cdot | s), \pie(\cdot | s)  \right) ]. \label{eq:param-objective} 
% % w_{t+1} &= \argmin_{w \in W} \mathop{\mathbb{E}}_{s \sim d^{\pi_{_{t}}}} [ D\left( \pi_w(\cdot | s), \pi_{t+1}(\cdot | s)  \right) ].
% \end{align}
% Unlike previous works, this update is compatible with continuous state-action spaces. For the remainder of the theory discussed below we will assume these distributions are over set of finite set of states and actions to ease notation. 
\begin{algorithm}
    \caption{Online optimization for OIL}
    \label{alg:FTL}
    \begin{algorithmic}[1]
        \State \textbf{Input:} Policy parameterization $\pi_{\x}$, Initial policy $\pi_{w_0}$ \Comment{$\pi_{w_0}$ has full support over the expert actions.}
        \State  \textbf{for} $t=1,2 ... T$ do
            \State \medspace \medspace \medspace \medspace Starting from $s \sim \rho$, roll-out policy $\pi_{\xt}$. 
            \Comment{Such interactions can be batched together.}
            \State  \medspace \medspace \medspace \medspace Construct the loss $l_t(\x)$ defined in~\cref{eq:pi-objective}.
        \State \medspace \medspace \medspace \medspace Update $\xtt$ using an online optimization algorithm. \Comment{e.g. OGD, AdaGrad, FTRL, FTL.}
        {
        \setlength{\abovedisplayskip}{2.5pt}
        \setlength{\belowdisplayskip}{2.5pt}
        }
        \State \textbf{end for} 
        \State \textbf{Output:} Control policy parameters $w_T$
    \end{algorithmic}  
\end{algorithm}
Notably, if the divergence is convex with respect to the parameterization $\x$, then each \emph{loss function} $\lt$ is also convex in $\x$, and OIL can be recast as an \emph{online convex optimization problem}~\citep{hazan2021introduction}. Note that here, the functions $l_t$ are not independent and identically distributed, but instead are generated by a complex interaction between the policy $\pit$ and the MDP at every iteration. In online optimization, we are tasked with finding a sequence of policies parameterized by $\{\x_1, \x_2, \ldots \x_T\}$ that minimize the sequence of loss functions $\lt$. Given $T$, the performance of an online optimization algorithm that produces a sequence $\{\x_1, \x_2, \ldots \x_T\}$ is measured in terms of its regret $R(T)$ defined as:
\begin{align} 
R(T) := \sum_{t=1}^{T} \lt(\xt) - \mathop{\min}_{\x \in \cW} \left[\sum_{t=1}^T \lt(\x) \right].
\label{eq:regret-def}
\end{align}
$R(T)$ measures the sub-optimality of the algorithm compared to the best performance in hindsight. We define $\cW^*_T := \argmin_{\x \in \cW}\sum_{t=1}^T l_t(\x) $ as the best parameter in hindsight. Algorithms that achieve sublinear regret for which $\lim_{T \rightarrow \infty} R(T)/T = 0$ are referred to as \emph{no-regret algorithms}. It is important to note that common algorithms~\citep{orabona2019modern} achieving sublinear regret do so for any (potentially adversarial) sequence of loss functions. Since it is difficult to correctly account for the interdependence of loss function and policy (which is to say, the policy plays a role in the generation of the next observed loss function), no-regret algorithms instead guarantee performance by safeguarding against worst-case behavior. In the next section, we focus on one such no-regret algorithm, follow-the-leader (FTL), and analyze its performance in OIL.

%% file: ftl.tex
\section{Follow-the-Leader }
\label{sec:ftl}
In this section, we begin by stating the Follow-the-Leader (FTL) update and then characterize its theoretical performance on OIL problems. The basic FTL update is given by:
\begin{equation}
\xtt \!\!=\!  \argmin_{\x \in \cW} \Ft(\x) := \sum_{i = 1}^t l_i(\x),
\label{eq:ftl-update}    
\end{equation} 
%The DAGGER algorithm~\citep{Ross2011} for example uses FTL to update the policy.
\!\!\!\! where $l_i(\x)$ is defined following~\cref{eq:pi-objective}. This algorithm is desirable because it is parameter free, and makes use of ``offline" updates~\citep{schulman2017proximal,mnih2016asynchronous,lillicrap2016continuous,schulman2015trust, degris2013offpolicy} by taking advantage of examples gathered during all previous interactions. These offline updates allow the algorithm to improve the policy without further interactions from the environment, and is important in settings where gathering environment interactions is expensive. In the general OCO framework, when the loss functions $\lt$ are strongly-convex in $\x$, FTL achieves $O\left(\log(T) \right)$ regret, but will incur $\Omega(T)$ regret in the absence of strong-convexity~\citep{hazan2007logarithmic}. However, these results do not capture the empirical success of FTL (e.g. DAGGER) used in conjunction with complex policy parameterizations like neural networks for which the loss functions are non-convex in $\x$~\citep{pmlr-v139-warrington21a}. 

To make progress towards addressing this discrepancy, we assume that (i) the policy class $\Pi$ is sufficiently expressive so as to contain the expert policy, (ii) the optimization problem in~\cref{eq:ftl-update} can be solved exactly, and (iii) the divergence $D(\pi, \pie)$ has a unique minimizer and is bounded in the sufficient statistics of $\pi$. Crucially, \textit{we do not make any assumptions about the policy parameterization}. Assumption (i) is true when using expressive policy classes like neural networks, while assumption (ii) relates to the supervised learning problem in~\cref{eq:ftl-update}. (ii) can be satisfied if the objective satisfies a gradient domination condition in $\x$, like for example, the PL inequality~\citep{karimi2016linear}. Lastly, assumption (iii) is typically true for the divergence-distribution pairs used in practice. For example, consider a continuous state-action space, where for a fixed state $s$ and $\pi(a|s) = \mathcal{N}(a; \mu^\top s, I)$ and $\pi_e(a|s) = \mathcal{N}(a; \mu_e^\top s, I)$, then $D(\pi, \pi_e) = \nicefrac{1}{2} \Vert \mu^\top s - \mu_e^\top s\Vert{}^2$. Under these assumptions, we prove (in~\cref{app:ftl-proofs}) FTL incurs constant regret.   
\newpage
\begin{restatable}[Follow-the-leader - Online Imitation Learning ]{theorem}{ftloil}
Under the following assumptions: (i) the policy class $\Pi$ is sufficiently expressive so as to contain the expert policy, the (ii) optimization problem in~\cref{eq:ftl-update} can be solved exactly, and (iii) the divergence $D(\pi, \pie)$ has a unique minimizer and is bounded in the sufficient statistics of $\pi$, FTL (\cref{eq:ftl-update}) obtains the following regret guarantee in the OIL setting,
\begin{equation*}
R(T) \leq \frac{C}{1-\gamma} \quad \text{where, } C := \max_{\tau,t} \left \{\mathop{\mathbb{E}}_{s\sim p^\tau_{t}(s)} [D(\pi(\cdot|s),\pi_e(\cdot|s))] \right \} < \max_{s} D(\pi(\cdot|s),\pi_e(\cdot|s)). 
\end{equation*} 
\!\! where $p^{\tau}_t(s)$ is the probability of reaching state $s$ at time-step $\tau$ using policy $\pit$. 
\label{thm:ftl}
\end{restatable} % here
The above result implies that FTL can take advantage of an expressive policy class, and obtain \emph{constant regret} in the OIL setting. Unlike the general online convex optimization results that require strong-convexity in $\x$ and imply a logarithmic regret for FTL (for completeness, we include these proofs in~\cref{app:ftlregret-sc}-\ref{app:ftlregret-scns}), the above theorem doesn't require the strong-convexity of $\lt(\x)$ and is in fact independent of parameterization. Next, we consider the practical implementation of FTL and discuss the advantages of using expressive policy classes in conjunction with FTL updates. 

%and since there is a one to one mapping from $\pi :\rightarrow \x$ for all $\pi \in \Pi$
\paragraph{Solving subproblem in~\cref{eq:ftl-update}:} If $\pie \in \Pi$, then there exists a $\x_e$ s.t. $\pie = \pi(\x_e)$. Since $D(\pi_e, \pi(\x_e)) = 0$, for all $t$, $\lt(\x_e) = 0$. Hence, the finite-sum problem in~\cref{eq:ftl-update} satisfies the \emph{interpolation}~\citep{vaswani2019fast,ma2018power} property. In this case, stochastic gradient descent (using a randomly sampled $l_i$) matches the convergence rate of deterministic gradient descent on $\Ft$. For example, if $\Ft(\x)$ satisfies the PL property, then it can be minimized to an $\epsilon$-error in $O(\log\left(\nicefrac{1}{\epsilon}\right))$ gradient evaluations, making the cost of the FTL update independent of $t$. We note that modern machine learning models (e.g. deep neural networks) used for OIL are sufficiently expressive and can ensure that the expert policy is contained in the resulting policy class. Furthermore, note that under interpolation, we can solve the subproblem using SGD with a stochastic line-search~\citep{vaswani2019painless}, making the FTL update  fully ``parameter-free''.       

We thus see that having an expressive policy class has a statistical (smaller number of interactions with the environment) as well as a computational advantage (small number of iterations to solve the  sub-problem for each update). But in cases where $\Pi$ does not contain $\pi_e$, we have demonstrated (see~\cref{fig:adversarial_grid_plots}) that FTL can result in linear regret and poor empirical performance. In order to remedy this issue, we consider a more general class of algorithms known as follow-the regularized-leader (FTRL)~\citep{abernethy2009competing}.

%% file: ftrl.tex
 \section{Follow-the Regularized-Leader}
\label{sec:ftrl}
Recall that for an expert with noisy feedback, FTL can lead to oscillations resulting in large cumulative regret (\cref{fig:adversarial_grid_plots}). We propose to use regularization to stabilize the behavior of FTL. In particular, we analyze the Follow-the Regularized-Leader (FTRL) algorithm. We first state the FTRL update and then reformulate it for a more scalable practical implementation. For smooth, convex losses, we quantify the regret of FTRL and its adaptive variant in~\cref{thm:ftrl-regret} and~\cref{thm:adaftrl-regret} respectively. The FTRL update (specifically the proximal variant given by~\citet{abernethy2009competing}) can be defined as: 
\begin{equation} \label{eq:ftrl-update}
\xtt \!\!=\!  \argmin_{\x \in \cW} \Ft(\x) + \psit(\x) := \left[\sum_{i = 1}^t l_i(\x) + \!\sum_{i = 1}^t \frac{\sigma_i}{2} \normsq{\x - \x_i} \right].  
\end{equation}
Note that FTRL can be used in conjunction with other regularizers~\citep{orabona2019modern}, but we focus on the squared Euclidean distance throughout this paper. The above update reduces to FTL (\cref{eq:ftl-update}) when $\sigma_i = 0$ for all $i$.~\cref{eq:ftrl-update}. Our analysis uses a proximal regularization term similar to~\citet{mcmahan2017survey}, though other variants also exist. Note that a naive implementation of~\cref{eq:ftrl-update} requires storing all the previous parameters ($\x_{1}, \x_{2}, \ldots, \xt$). This issue is exacerbated when using large, complex models to parameterize the policy. Using FTRL for continual learning also results in the same problem~\citep{kirkpatrick2017overcoming}, and is tackled heuristically. Instead, we reformulate the update in~\cref{eq:ftrl-update} as follows:
\begin{restatable}[Reformulation]{proposition}{ftrlreformulation}
Defining $\etat := \nicefrac{1}{\left(\sum_{i=1}^t \sigma_i\right)}$, the update in~\cref{eq:ftrl-update} can be reformulated (proof in~\cref{app:ftrl-proofs}) as: 
\begin{align}
\xtt = \argmin_{\x \in \cW} \left[
\sum_{i = 1}^t l_i(\x) - \inner{\x}{\sum_{i = 1}^{t-1} \nabla l_i(\xt)} + \frac{1}{2 \etat} \normsq{\x - \xt} \right]. \label{eq:ftrl-reformulation}
\end{align}
\label{prop:ftrl-reformulation}
\end{restatable} 
Unlike~\cref{eq:ftrl-update}, this update does not have a memory requirement which increases with the number of iterations and model size. That is, if $m$ is the model-size, then~\cref{eq:ftrl-update} requires $O(mT)$ memory, while the reformulated update can be implemented using only $O(m + T)$ memory (same as FTL).  We note such reformulations are not unique, and choosing one reformulation over another could lead to drastically different solutions in settings where the inner optimization problem defined by \eqref{eq:ftrl-update} is non-convex, or solved inexactly. While we leave a theoretical discussion on this topic to future work, we include an additional reformulation for comparison which we refer to as Alt-FTRL, 
\begin{align}
\xtt = \argmin_{\x \in \cW} \left[
\sum_{i = 1}^t l_i(\x) + 
\frac{1}{2 \etat} \normsq{\x} 
- \x^\top \left[\sum_{i = 1}^{t-1} \xt \left[ \frac{1}{\eta_t} - \frac{1}{\eta_{t-1}}\right]\right]\right]. \label{eq:alt-ftrl-reformulation}
\end{align}  
This reformulation averages over the previous parameters instead of their gradients, and is included in the empirical comparison in \cref{sec:experiments}. Next, we describe how interpolation improves the computational efficiency of FTRL.   
\paragraph{Solving subproblem in~\cref{eq:ftrl-update}:} Similar to~\cref{eq:ftl-update}, if $\pie \in \Pi$, the finite-sum in $\Ft(\x)$ satisfies the interpolation property defined in~\cref{sec:ftl}. In this case, proximal stochastic gradient descent (using a randomly sampled $l_i$ and a proximal operator with $\psi_t(\x)$ discussed in \cref{app:ftrl-proofs}) matches the convergence rate of deterministic gradient descent~\citep{cevher2019linear}. Hence, similar to~\cref{eq:ftl-update},~\cref{eq:ftrl-update} can be solved efficiently. Unlike the result in~\cref{sec:ftl}, here, we assume that the loss functions are $L$-smooth and convex in $\x$  (see~\cref{app:definitions} for formal definitions and \cref{app:ftrl-nonsmooth} for proofs under the convex, non-smooth but Lipschitz setting). The subsequent results also assume that $\cW$ is a convex compact set of diameter $D$, meaning $\sup_{x, y \in X}\norm{x - y} \leq D$. We use the following definition to quantify the degree to which the  interpolation property is satisfied. 
\begin{definition}[Interpolation-Error]
For a fixed iteration $t$, if $\xopt_t := \argmin_{\x \in \cW} \lt(\x)$, and for a $\xopt \in \cW^*$,  
\begin{equation}
\epsilon_t^2 := \min_{\xopt \in \cW^*} \{ l_t(\xopt) - l_t(\xopt_t) \}.
\end{equation}
\end{definition} 
If $\pie \in \Pi$, then $\x_e \in \cW^*$. We know that $l_t(\x_e) = 0$ for all $t$, and since each $\lt$ is lower-bounded by zero, it implies that $\epsilon^2_t = 0$ for all $t$. Hence, $\epsilon^2_t$ is a measure of the expressivity of the policy class that is induced by the $\x :\rightarrow \pi$ mapping and the set $\cW$. In the following theorem (proved in~\cref{app:ftrl-proofs}), under the above assumptions, we show that FTRL incurs sublinear regret regardless of whether $\pie \in \Pi$.   
\begin{restatable}[FTRL - Smooth + Convex]{theorem}{ftrlregret}
Assuming each $l_t$ is (i) $L$-smooth, (ii) convex, FTRL (\cref{eq:ftrl-reformulation}) for $\eta_t = \min \left\{(\sum_{t = 1}^T \epsilon_t^2)^{-1/2}, \frac{1}{2L} \right\}$ for all $t$, achieves the following regret 
\begin{align*}
R(T) &\leq \sum_{t=1}^T \left[ \frac{\etat \normsq{\nabla \lt(\x_t)}}{2} \right] + \frac{D^2}{2 \eta_{T}} \leq 2D^2L + (D^2+2L) \sqrt{\sum_{t=1}^{T} \epsilon^2_t}. 
\end{align*}
\label{thm:ftrl-regret}
\end{restatable} 
This result follows similar proof techniques established in works like~\citet{orabona2019modern,ghadimi2013stochastic}, however unlike previous results for FTRL, we present a bound which explicitly accounts for the level of interpolation similar to~\citep{loizou2020stochastic,vaswani2020adaptive}. If $\pie \in \Pi$, $\epsilon_t = 0$ for all $t$, and FTRL achieves constant regret similar to FTL. For non-zero $\epsilon_t$, FTRL still incurs sublinear $O(\sqrt{T})$ regret. However, we note that unlike~\cref{thm:ftl}, the above result requires the loss functions to be convex and smooth. Unfortunately, the above result requires setting $\eta$ according to $L$ and $\epsilon_t$, both of which are typically unknown in practice. To address this issue, we use an adaptive variant of FTRL~\citep{joulani2020modular,mcmahan2017survey} and characterize its regret bound in the following theorem (proved in~\cref{app:ftrl-proofs}).   
\begin{restatable}[AdaFTRL - Smooth + Convex]{theorem}{adaftrlregret}
Assuming each $l_t$ is (i) $L$-smooth, (ii) convex, FTRL (\cref{eq:ftrl-reformulation}) for $\eta_t = \nicefrac{\alpha}{\sqrt{\sum_{i = 1}^t ||\nabla l_i(\x_i) ||^2}}$, achieves the following regret  
\begin{align*}
R(T) &\leq \sum_{t=1}^T \left[ \frac{\alpha ||\nabla l_t(\x_t)||^2}{2\sum_{i=1}^{t}||\nabla l_i(\x_i)||^2} \right] + \frac{D^2}{2\alpha} \sqrt{\sum_{t=1}^{T} ||\nabla l_t(\x_t)||^2} \leq 2L \left(\frac{\alpha}{2} + \frac{D^2}{2\alpha} \right)^2 + \sqrt{2L} \left(\frac{\alpha}{2} + \frac{D^2}{2\alpha} \right) \sqrt{\sum_{t=1}^{T} \epsilon^2_t}.
\end{align*}
\label{thm:adaftrl-regret}
 \end{restatable} 
Observe that the above bound holds for any finite value of $\alpha$, though the upper-bound is minimized when $\alpha = D$. The only difference with the FTRL update used in~\cref{thm:ftrl-regret} is the choice of $\etat$. Hence, we can continue to use the reformulated update in~\cref{prop:ftrl-reformulation}. Furthermore, in smooth convex settings, AdaFTRL achieves the same regret as in~\cref{thm:ftrl-regret}  without the knowledge of problem-dependent constants like $L$ or $\epsilon_t$. The proof of~\cref{thm:adaftrl-regret} is similar to AdaGrad~\citep{vaswani2020adaptive,levy2018online}. We conclude by showing that FTRL (and AdaFTRL) generalizes OGD (and AdaGrad) respectively. In particular, if we were to use the linearized losses, meaning that $l_s(\x) = \langle \nabla l_s(\xs), \x \rangle$ in~\cref{prop:ftrl-reformulation}, then, by definition of $\xtt$, setting the gradient to zero, $\sum_{i = 1}^t \nabla l_i(\x_i) - \sum_{i = 1}^{t-1} \nabla l_i(\x_i) + \etat (\xtt - \xt) = 0, \implies \xtt = \xt - \etat \nabla \lt(\xt)$, which recovers the OGD (and AdaGrad) update with the corresponding choice of $\etat$.

%% file: experiments.tex
\section{Experiments}
\label{sec:experiments}
In this section, we compare FTL, FTRL,   Alt-FTRL, AdaFTRL, online gradient descent~\citep{zinkevich2003online} (OGD) and AdaGrad~\citep{duchi2011adaptive} in terms of both the average cumulative loss   equal to $\nicefrac{1}{t}\sum_{s = 1}^{t} l_s(\x_s)$ and the policy return $V^\pit(\rho)$. Every round consists of $M$ interactions with the environment and we evaluate each algorithm for a total of $25000$ ($10000$ for Atari) environment interactions, meaning that  $T = \nicefrac{25000}{M}$. Throughout the main paper, we use $M = 1000$, and defer the results for $M = 100$ to~\cref{app:exp-details}). In order to tune the "outer-learning-rate" $\eta$ for OGD, AdaGrad, and the FTRL variants, we do a grid-search over $\eta \in [10^{-5}, 10^{-4}, \ldots, 10^{5}]$ on one of the environments, and take the top three step-sizes that has the minimum average cumulative loss over the course of $2000$ environment interactions under $M=100$. We then evaluate these three step-sizes over 25000 interactions, and take the best of the three in terms of average cumulative loss. For the off-policy methods (FTL and the FTRL variants), we also search over the space of "inner-learning-rates" $\alpha \in [10^{-5}, 10^{-4}, \ldots, 10^{5}]$, and run two optimization procedures, Adam~\citep{kingma2014adam}, and SLS~\citep{loizou2020stochastic}. For FTRL and Alt-FTRL, we set $\eta_t = \nicefrac{\alpha}{\sqrt{t}}$. For FTL, FTRL and its variants, we used gradient descent to solve the subproblems for each update. For solving each subproblem, we used a maximum of $1000$ iterations terminating the optimization when the gradient norm was sufficiently small ($10^{-8}$).  

\subsection{Continuous Control on Mujoco} 
\label{sec:exps-mujoco}
We evaluate the algorithms for continuous control tasks (with continuous state and action spaces) in Mujoco suite~\citep{todorov2012mujoco}, and build and train models using pytorch~\citep{pytorch}. In particular, we consider the Hopper and Walker-2D environments where the task is to learn a policy that can imitate the expert policy trained used reinforcement learning. The expert policy uses a neural network parameterization and is trained using soft actor-critic~\citep{haarnoja2018soft}. For each environment, we report the performance of each method when the loss function $\lt(\x)$ is either the $l_2$ or $l_1$ loss. All results are averaged over $3$ runs, and we report the mean and relevant quantiles. The policy corresponds to a multivariate Gaussian distribution with a fixed diagonal covariance and the mean parameterized by either a linear or neural network model. The neural network architecture is the same as that of the expert, meaning that in this case, the policy class is sufficiently expressive to include the expert policy. For the linear model, the resulting loss functions are convex, whereas using a neural network parameterization results in non-convex loss functions. Because of its poor empirical performance, we do not plot the OGD in the main paper and defer these plots to~\cref{app:exp-details}. 

\begin{figure}[t]
    \centering
    \includegraphics[width=\textwidth]{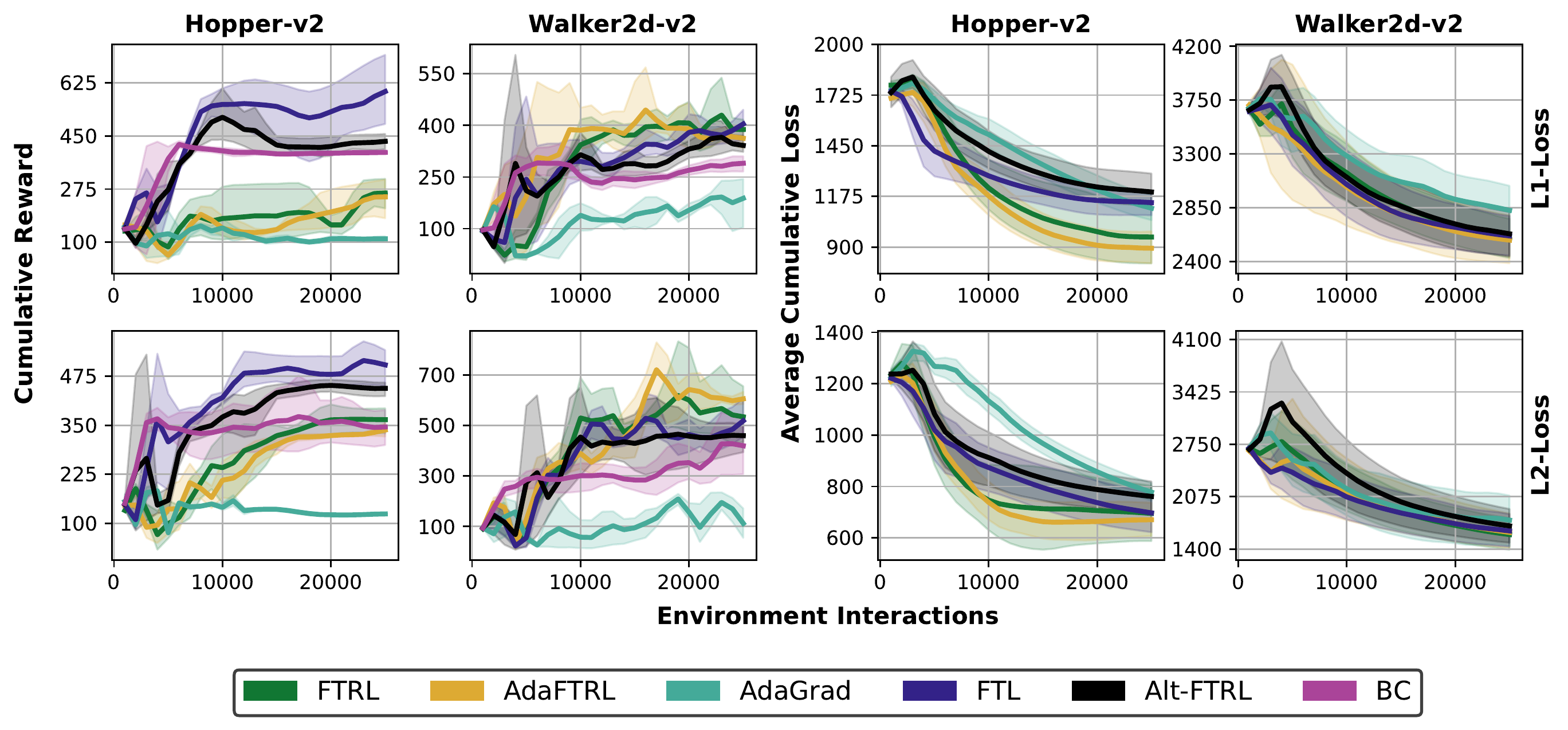}
    \caption{Comparing Adagrad, FTL, FTRL, Alt-FTRL, and AdaFTRL, the four plots on the left display the expected cumulative reward while average cumulative loss displayed on the right. A behavioral cloning baseline (BC) is included to verify the effect of environment interaction under the learned policy. Each line describes the mean, $5\%$ and $95\%$ quantiles as computed by~\citet{2020NumPy-Array}. The plots for the average cumulative loss show that for \textbf{linear models}, (i) FTRL, Alt-FTRL and AdaFTRL maintain performance that is as good or better than FTL in terms of average cumulative loss, indicating the benefits of regularization, and (ii) FTL, FTRL, and AdaFTRL do significantly better than AdaGrad, indicating that making ``offline" updates are crucial in maintaining performance in online imitation learning.}
    \label{fig3:linear}
\end{figure}

\begin{figure}[t]
    \centering
    \includegraphics[width=\textwidth]{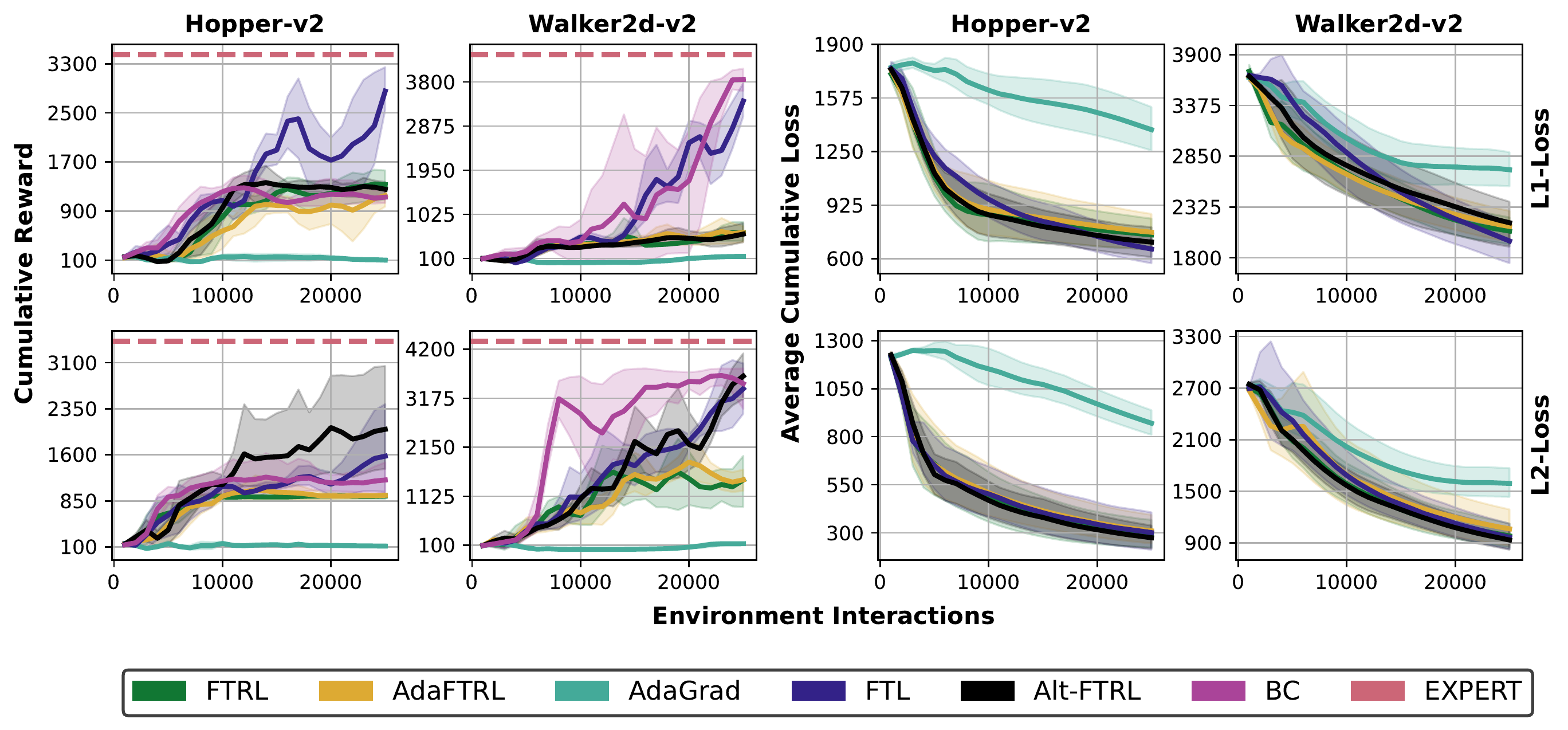}
    \caption{Comparing Adagrad, FTL, FTRL, Alt-FTRL and AdaFTRL, the four plots on the left display the expected cumulative reward in left 4 plots and Average cumulative loss on the right four plots. A behavioral cloning baseline (BC) is included to verify the effect of environment interaction under the learned policy, and expert performance is indicated with a dotted line.  Each line presents the mean, $5\%$ and $95\%$ quantiles as computed by \citet{2020NumPy-Array}. These plots show that for \textbf{neural network models}, (i) FTRL, Alt-FTRL, and AdaFTRL maintain performance in terms of cumulative loss that is as good as FTL, even under a complex model class,  and (ii) FTL, FTRL, and AdaFTRL all do significantly better than their on-policy counterparts, indicating that making ``offline" updates can be crucial in maintaining performance.}
    \label{fig3:nn} 
\end{figure}

For both parameterizations (\cref{fig3:linear} and~\cref{fig3:nn}), we observe that (i) FTL, FTRL, Alt-FTRL, and AdaFTRL consistently outperform OGD and AdaGrad, (ii) FTRL and its variants consistently outperform FTL in terms of the average cumulative loss, (iii) AdaFTRL improves over both FTRL and Alt-FTRL in terms of average cumulative loss,  (iv) FTL has good performance for the non-strongly-convex $l_1$ loss in the linear case where interpolation is not necessarily satisfied, and (v) good performance with respect to the average loss metric does not imply good return (for example, with the linear model and L2 loss, AdaGrad matches the average loss of the other methods, but has poor performance with respect to the cumulative return). We conclude that (i) ``offline'' updates used in FTL, FTRL and its variants result in superior empirical performance and (ii) regularization helps improve the empirical performance with FTRL outperforming FTL in terms of average cumulative loss but not always average cumulative reward, and (iii) FTL performs better compared to what is suggested by the theory. 

\subsection{Discrete Control on Atari} 
\label{sec:exps-atari}
We evaluate the algorithms for discrete control tasks (with discrete state-action spaces) in the Atari suite~\citep{mnih2013playing}. In particular, we consider the Pong and Breakout game environments where the task is to learn a policy that can imitate the expert policy trained used reinforcement learning. The expert policy uses a neural network parameterization and is trained using proximal policy optimization~\citep{stable-baselines,schulman2017proximal}. The learned policy corresponds to a categorical distribution parameterized by either a linear model that uses the fixed pretrained features from the reinforcement learning algorithm, or the same neural network architecture as the expert and is learned in an end-to-end fashion. For the linear model (which uses the pretrained feature extractor from the expert), the resulting loss functions are convex, while the end-to-end setup tests the non-convex setting. Because of its poor empirical performance, we again do not plot OGD in the main paper and defer these plots to~\cref{app:exp-details}. Notably both settings the policy class is sufficiently expressive so as to include the expert policy.

\begin{figure}[t]
    \centering 
    \includegraphics[width=\textwidth]{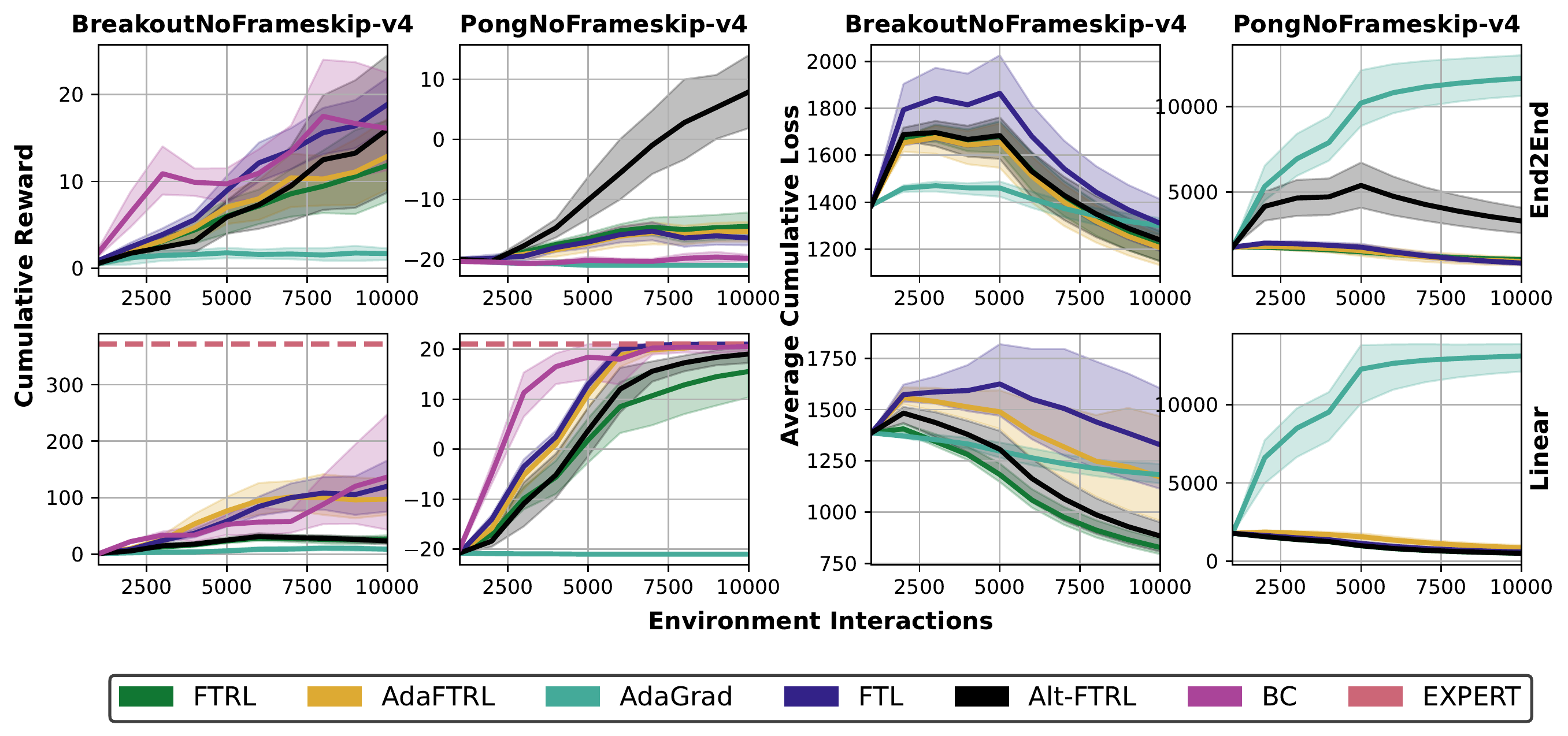} 
    \caption{Comparing Adagrad, FTL, FTRL, Alt-FTRL, and AdaFTRL, we plot the the expected cumulative reward on the left 4 plots and the average cumulative loss on the right. Each line presents the mean, $5\%$ and $95\%$ quantiles as computed by \citet{2020NumPy-Array}. We observe that for the end-to-end setting, (i) FTRL, Alt-FTRL, and AdaFTRL outperform FTL in terms of average cumulative loss, and (ii) FTL, FTRL, Alt-FTRL, and AdaFTRL all do significantly better than AdaGrad, indicating that ``offline" updates can be crucial for good empirical performance. A behavioral cloning baseline (BC) is included to verify the effect of environment interaction under the policy, and expert performance is indicated with a dotted line. We also note that the "linear policy class" represents a linear transformation of the experts pretrained feature encoder. This ensures that the learned policy satisfies the interpolation convexity assumptions.}
    \label{fig3:atari} 
\end{figure}

In~\cref{fig3:atari}, we again observe that for both policy parameterizations, (i) FTL, FTRL and AdaFTRL consistently outperform OGD and AdaGrad, (ii) FTRL and AdaFTRL often dominate FTL in terms of the average cumulative loss, (iii) AdaFTRL has similar performance as its non-adaptive variants in terms of the average cumulative loss, (iv) FTL has good performance for the non-strongly-convex cross-entropy loss, and (v) similar to~\cref{sec:exps-mujoco}, good performance with respect to the average loss metric does not imply good return. We again conclude that (i) ``offline'' updates used in FTL, FTRL and its variants result in superior empirical performance and (ii) regularization helps improve the empirical performance with FTRL outperforming (in average cumulative loss) FTL in the end-to-end setting, and (iii) FTL performs better compared to that suggested by the theory. 

%% file: related-work.tex
\section{Other Related Work} 
\label{sec:related-work}
We briefly discussed the most relevant related work in~\cref{sec:introduction}. We now clarify how OIL relates to other common settings, focusing specifically on imitation learning from observation alone or ILOA~\citep{DBLP:journals/corr/abs-2007-02520, sun2019provably}. Unlike OIL, ILOA often requires solving difficult sub-problems at every round of interaction. One of the most popular variants of ILOA called \textit{apprenticeship learning - AL}~\citep{shani2021online,zahavy2020apprenticeship,abbeel2004apprenticeship}, uses the state-occupancy generated by the expert to construct a reward surface, and solves for the optimal policy under this surface. For example, \citet{syed2007game} assume (a) access to expert trajectories and that (b) the underlying unknown reward function is a non-negative linear combination of known "state-features", and solve a saddle-point problem by alternating between 1) constructing a prospective reward under which the expert occupancy is optimal and the agent's occupancy is maximally sub-optimal, and 2) using the prospective reward, solve an RL problem via policy optimization. In contrast, the basic OIL setting~\citep{ross2011reduction,DBLP:journals/corr/abs-2007-02520} make no assumption about the unknown reward surface, and instead assumes access to an expert oracle which provides the optimal action given a state. Here, the OIL problem is typically framed as an online optimization problem, can be solved by standard OCO techniques, and notably doesn't require solving RL problems as a subroutine.

A variety of authors improve upon the framework given by~\citep{syed2007game, abbeel2004apprenticeship}, often taking advantage of theoretical advances in constrained optimization. For example, Frank-Wolfe updates~\citep{zahavy2020apprenticeship, abernethy2017frank}, can be used to solve classical variants of the apprenticeship learning problem, and results which extend this framework have even proven some convergence results in non-stationary tabular MDPs~\citep{geist2021concave, zahavy2021reward,zhang2020variational}. In contrast to these works, our results in Section 4 can handle more complex policy parameterizations (e.g. linear) given that the corresponding losses satisfy the appropriate convexity assumptions, while the FTL results in Section 3 can handle general policy parameterizations under an interpolation assumption (extending similar work in the bandits setting by \citet{degenne2018bandits}). In some cases by framing the saddle point problem as online mirror-decent~\citep{shani2021online}, one can convert the computationally costly saddle-point problem into a more tractable iterative algorithm similar to \textit{adversarial imitation learning} (AIL)~\citep{creswell2018generative, ghasemipour2020divergence, fu2018learning}. Again however, these algorithms require stricter assumptions on the class of MDPs considered, or have no guarantees at all. From a sample-complexity perspective, \citet{pmlr-v70-baram17a} shows by assuming access to an expert oracle (like in OIL), the number of environment interactions required to match the expert performance $\approx 10^5$, while AIL-like algorithms require up to $10^6$ even with a similar number of expert examples. Addressing this gap in statistical efficiency between these algorithm classes, and extending results based non-stationary MDPs to continuous state-action spaces represent interesting areas of future research.

%% file: discussion.tex
\section{Discussion} 
\label{sec:conclusion}
We show that in OIL settings (i) algorithms which make use of offline updates (FTL, FTRL, AltFTRL, AdaFTRL) perform better than algorithms which do not (OGD, AdaGrad) and (ii) including regularization can lead to empirical improvements in terms of reward and average cumulative loss. Furthermore, we improved the theoretical results for both FTL and FTRL, showing that both algorithms can achieve constant regret when the policy class is sufficiently expressive and contains the expert policy. Importantly, our guarantees for FTL only require that the losses be strongly-convex with respect to the policy's sufficient statistics (not its parameters). Our research leaves a host of open questions -- (i) does FTRL converge for the IL setting independent of parameterization in a similar fashion to FTL and (ii) can these results be generalized to the standard online learning setting without leveraging OIL structure, and (iii) how do inexact optimization, stochastic gradients, and limited memory affect performance of FTL and FTRL.

%% file: acknowledgements.tex
\section{Acknowledgements}
We thank the anonymous reviewers, whose comments helped improve the clarity of the manuscript. We thank Frederik Kunstner for his invaluable discussion, for providing comments on the manuscript and earlier versions of this work, and for suggesting related material. We also thank Victor Sanches Portella and Betty Shea for additional discussions we believe improved the overall clarity of the work.  This research was partially supported by the Canada CIFAR AI Chair Program, the Natural Sciences and Engineering Research Council of Canada (NSERC) Discovery Grants RGPIN-2015-06068 and the NSERC Post-graduate Scholarships-Doctoral Fellowship 545847-2020.

%% file: app-setup.tex
\newcommand{\appendixTitle}{%
\vbox{
    \centering
	\hrule height 4pt
	\vskip 0.2in
	{\LARGE \bf Supplementary material}
	\vskip 0.2in
	\hrule height 1pt 
}}

\appendixTitle

\section*{Organization of the Appendix}
\begin{itemize}

 \item[\ref{app:definitions}] \nameref{app:definitions}
 
 \item[\ref{app:ftl-proofs}] \nameref{app:ftl-proofs}
 
 \item[\ref{app:ftrl-proofs}] \nameref{app:ftrl-proofs}
 
 \item[\ref{app:exp-details}] \nameref{app:exp-details}

\end{itemize}

\section{Definitions}
\label{app:definitions}
Our main assumptions are that each individual function $l_i$ is differentiable, has a finite minimum $l_i^*$, and is $L$-smooth, meaning that for all $v$ and $w$, 
\aligns{
    l_i(v) & \leq l_i(w) + \inner{\nabla l_i(w)}{v - w} + \frac{L}{2} \normsq{v - w},
    \tag{Individual Smoothness}
    \label{eq:individual-smoothness}
}
which also implies that $f$ is $L$-smooth. A consequence of smoothness is the following bound on the norm of the stochastic gradients,
\aligns{
    \norm{\nabla l_i(\x)}^2 
    \leq
    2 L (l_i(\x) - l_i^*).
}
We also assume that each $l_i$ is convex, meaning that for all $v$ and $w$,
\aligns{
    l_i(v) &\geq l_i(w) - \langle \nabla l_i(w), w-v \rangle,
    \tag{Convexity}
    \label{eq:individual-convexity}
    \\
}
Depending on the setting, we will also assume that $f$ is $\mu$ strongly-convex, meaning that for all $v$ and $\x$,
\aligns{
f(v) & \geq f(w) + \inner{\nabla f(w)}{v - w} + \frac{\mu}{2} \normsq{v - w},
\tag{Strong Convexity}
\label{eq:strong-convexity}
}

%% file: app-proofs-ftrl.tex
\section{Proofs for Section~\ref{sec:ftrl}}
\label{app:ftrl-proofs}

\subsection{Proof of Proposition~\ref{prop:ftrl-reformulation}}

\begin{thmbox}
\ftrlreformulation*
\end{thmbox}
\begin{proof}
Since $\etat := \frac{1}{\sum_{s = 1}^{t} \sigmas}$, by definition of $\xtt$ in~\cref{eq:ftrl-update},
\begin{align*}
\sum_{s = 1}^{t} \grads{\xkk} + \frac{\xtt}{\etat} = \sum_{s = 1}^{t} \sigmas \xs
\end{align*}
Similarly, by definition of $\xt$, 
\begin{align*}
\sum_{s = 1}^{t-1} \grads{\xt} + \frac{\xt}{\eta_{t-1}} = \sum_{s = 1}^{t-1} \sigmas \xs
\end{align*}
From the above relations, 
\begin{align*}
\sum_{s = 1}^{t} \grads{\xtt} + \frac{\xtt}{\etat} & = \sum_{s = 1}^{t-1} \grads{\xt} + \sigmak \xt + \frac{\xt}{\eta_{t-1}} \\
\implies \xtt + \etat \sum_{s = 1}^{t} \grads{\xtt}  & = \xt + \etat \sum_{s = 1}^{t-1} \grads{\xt} 
\end{align*}
Therefore, at iteration $t$, we need to obtain $\xtt$, we need to solve the following equation w.r.t $\x$,
\begin{align}
\x + \etat \sum_{s = 1}^{t} \grads{\x}  & = \xt + \etat \sum_{s = 1}^{t-1} \grads{\xt}
\label{eq:ftrlpractical}
\end{align}
Similar to FTL, this update requires storing the previous functions $f_s$ from $s = 1$ to $t$, but does not require storing the previous models like in~\cref{eq:ftrl-update}. Minimizing the following loss is equivalent to ensuring~\cref{eq:ftrlpractical}. 
\begin{align}
\xtt &= \argmin_{\x} \left[
\sum_{s = 1}^{t} l_s(\x) - \inner{\x}{\sum_{s = 1}^{t-1} \grads{\xk}} + \frac{1}{2 \etat} \normsq{\x - \xk}
\right]  
\label{eq:ftrl-loss}
\end{align}
\end{proof}

\subsection{Derivation of Alternative FTRL Reformulation in ~\eqref{eq:alt-ftrl-reformulation}} 
In a more direct fashion then the proof above, we can see that 
\begin{align}
\xtt = \argmin_{\x \in \cW} \left[
\sum_{i = 1}^t l_i(\x) + 
\frac{1}{2 \etat} \normsq{\x} 
- \sum_{i = 1}^{t-1} \x^\top\xt \left[ \frac{1}{\eta_t} - \frac{1}{\eta_{t-1}}\right]\right], 
\end{align}
In fact represents the same objective as
\begin{align}
\xtt = \argmin_{\x \in \cW}  
\left[\sum_{i = 1}^t l_i(\x) + \!\sum_{i = 1}^t \frac{\sigma_i}{2} \normsq{\x - \x_i} \right] 
\end{align}
We can see this if we differentiate both equations
\begin{align}
\nabla F_{\text{alt}}(\x) &= \nabla \left[
\sum_{i = 1}^t l_i(\x) + 
\frac{1}{2 \etat} \normsq{\x} 
- \sum_{i = 1}^{t-1} \x^\top\x_i \left[ \frac{1}{\eta_i} - \frac{1}{\eta_{i-1}}\right]\right] \nonumber  \\  
&= 
\nabla \left[ \sum_{i = 1}^t   l_i(\x) + 
   \frac{\x^\top\x}{2} \sum_{i = 1}^{t-1} \left[ \frac{1}{\eta_i} - \frac{1}{\eta_{i-1}}\right] 
- \sum_{i = 1}^{t-1} \x^\top\x_i \left[ \frac{1}{\eta_i} - \frac{1}{\eta_{i-1}}\right]\right] \nonumber  \\
&= 
\nabla \left[ \sum_{i = 1}^t   l_i(\x) + 
   \frac{\x^\top\x}{2} \sum_{i = 1}^{t-1} \left[ \frac{1}{\eta_i} - \frac{1}{\eta_{i-1}}\right] 
- \sum_{i = 1}^{t-1} \x^\top\x_i \left[ \frac{1}{\eta_i} - \frac{1}{\eta_{i-1}}\right] + 
   \sum_{i = 1}^{t-1} \frac{\x_i^\top\x_i}{2}  \left[ \frac{1}{\eta_i} - \frac{1}{\eta_{i-1}}\right] \right] \nonumber \\
 &= 
\nabla \left[ \sum_{i = 1}^t   l_i(\x) +
     \sum_{i = 1}^{t-1} \frac{\sigma_i}{2} \normsq{\x - \x_i} \right] \nonumber  \\
     &= \nabla F(\x)
\end{align}
Note that we define, 
\begin{equation}
    \sigma_t=\frac{1}{\eta_t} - \frac{1}{\eta_{t-1}},\quad \text{and}\quad \frac{1}{\eta_t}=\sqrt{t}
\end{equation}
to ensure that the magnitude of regularization used in both reformulations is of order $\sqrt{t}$. 
\subsection{Main Regret Lemma}
\label{app:ftrl-lemma}
We define 
\begin{align}
\Ft(\x) := \sum_{i = 0}^{t-1} \li(\x)  + \psit(\x),
\label{eq:F-def}
\end{align}
where, 
$\psit(\x)$ is a strongly-convex proximal regularizer that satisfies the following property:
\begin{align}
\xt &= \argmin \left[\psitt(\x) - \psit(\x) \right].     
\label{eq:prox-reg}
\end{align}
\cref{eq:ftrl-update} uses $\psit(\x) = \sum_{i = 1}^{t-1} \frac{\sigma_i}{2} \normsq{\x - \x_i}$. Since $\psitt(\x) - \psit(\x) = \frac{\sigma_t}{2} \normsq{\x - \x_t}$, which is minimized at $\x_t$ and hence the regularizer in~\cref{eq:ftrl-update} satisfies the desired property, and is a valid strongly-convex proximal regularizer. We will now prove the following lemma for a general strongly-convex proximal regularizer $\psit$. In this case, the FTRL update in~\cref{eq:ftrl-update} can be generalized to: 
\begin{align}
\xt = \argmin \Ft(\x) = \argmin \sum_{i = 0}^{t-1} \li(\x)  + \psit(\x) \implies \sum_{i = 1}^{t - 1} \gradi{\xt} + \nabla \psit(\xt) = 0. 
\label{eq:ftrl-update-gen}
\end{align}  
\begin{thmbox}
\begin{lemma}
Assuming that the functions $\Ft$ are $\lambdat$-strongly convex, then the regret for the FTRL update in~\cref{eq:ftrl-update-gen} can be bounded as:
\begin{align*}
R(T) \leq [F_1(\x_1)] +  \sum_{t = 1}^{T} \left[\frac{1}{2 \lambdatt} \normsq{\gradt{\xt}} \right] - \sum_{t = 1}^{T} [\psit(\xt) - \psitt(\xt)] + \psi_{T+1}(\xopt)     
\end{align*}
\label{lemma:ftrl-main}
\end{lemma}
\end{thmbox}
\begin{proof}
\begin{align*}
\Ftt(\xt) - \Ftt(\xtt)  & \leq \frac{1}{2 \lambdatt} \normsq{\nabla \Ftt(\xt)} & \tag{Since $\Ftt$ is $\lambdatt$-strongly convex and $\xtt$ is the minimizer of $\Ftt$.} \\
& = \frac{1}{2 \lambdatt} \normsq{\sum_{i = 1}^{t} \gradi{\xt} + \nabla \psitt(\xt)} & \tag{By definition of $\Ftt$} \\
& = \frac{1}{2 \lambdatt} \normsq{\sum_{i = 1}^{t-1} \gradi{\xt} + \nabla \psit(\xt)  + \nabla \psitt(\xt) - \nabla \psit(\xt) + \gradt{\xt}} \\
& = \frac{1}{2 \lambdatt} \normsq{\nabla \Ft(\xt) + \nabla \psitt(\xt) - \nabla \psit(\xt) + \gradt{\xt}} \tag{By definition of $\Ft$} \\
& = \frac{1}{2 \lambdatt} \normsq{\nabla \psitt(\xt) - \nabla \psit(\xt) + \gradt{\xt}} \tag{Since $\xt$ is the minimizer of $\Ft$} \\
& = \frac{1}{2 \lambdatt} \normsq{\gradt{\xt}} \tag{Since $\xt$ is the minimizer of $\psitt(\x) - \psit(\x)$} 
\end{align*}

\begin{align*}
\Ftt(\xt) - \Ftt(\xtt) & = [\Ftt(\xt) - \Ft(\xt)] + [\Ft(\xt) - \Ftt(\xtt)] \\
& = [\lt(\xt) + \psitt(\xt) - \psit(\xt)] + [\Ft(\xt) - \Ftt(\xtt)] \\
\end{align*}
Summing from $t = 1$ to $T$, and using the above relation,
\begin{align*}
& \sum_{t = 1}^{T} [\lt(\xt) + \psitt(\xt) - \psit(\xt)] + \sum_{t = 1}^{T} [\Ft(\xt) - \Ftt(\xtt)] \leq \sum_{t = 1}^{T} \left[\frac{1}{2 \lambdatt} \normsq{\gradt{\xt}} \right] \\
& \implies \sum_{t = 1}^{T} [\lt(\xt) - \lt(\xopt)] + [F_1(\x_1) - F_{T+1}(\x_{T+1})]  \leq \sum_{t = 1}^{T} \left[\frac{1}{2 \lambdatt} \normsq{\gradt{\xt}} \right] - \sum_{t = 1}^{T} [\psit(\xt) - \psitt(\xt)] - \sum_{t = 1}^{T} \lt(\xopt) \\
& R(T) \leq [F_{T+1}(\x_{T+1}) - F_{T+1}(\xopt) - F_1(\x_1)] +  \sum_{t = 1}^{T} \left[\frac{1}{2 \lambdatt} \normsq{\gradt{\xt}} \right] - \sum_{t = 1}^{T} [\psit(\xt) - \psitt(\xt)] + \psi_{T+1}(\xopt) \\
\intertext{Since $\x_{T+1}$ is the minimizer of $F_{T+1}$,}
& R(T) \leq [F_1(\x_1)] +  \sum_{t = 1}^{T} \left[\frac{1}{2 \lambdatt} \normsq{\gradt{\xt}} \right] - \sum_{t = 1}^{T} [\psit(\xt) - \psitt(\xt)] + \psi_{T+1}(\xopt) 
\end{align*}
\end{proof}

The above expression is true for both FTL, and (adaptive) FTRL, and only uses the definitions of the proximal regularizer and the strong-convexity property for $\Ft$. We specialize this result for $\psit(\x) = \sum_{i = 1}^{t-1} \frac{\sigma_i}{2} \normsq{\x - \x_i}$ used in~\cref{eq:ftrl-update}. 
\begin{thmbox}
\begin{lemma}
Assuming that each $\li$ is $\mu_i$ strongly-convex for $\mu_i \geq 0$, the regret for the FTRL update in~\cref{eq:ftrl-update} can be bounded as:
\begin{align*}
R(T) & \leq \sum_{t = 1}^{T} \left[\frac{1}{2 \sum_{i= 1}^{t} [\sigma_i + \mu_i]} \normsq{\gradt{\xt}} \right] + \frac{D^2}{2} \; \sum_{t = 1}^{T} \sigma_t
\end{align*}
\label{lemma:ftrl-main-l2}
where $D$ is the diameter of $\cW$. 
\end{lemma}
\end{thmbox}
\begin{proof}
With this choice of $\psit$, we note that $F_1(x_1) = \psi_1(x) = 0$ for all $\x$. Using~\cref{lemma:ftrl-main}, 
\begin{align*}
R(T) &  \leq \sum_{t = 1}^{T} \left[\frac{1}{2 \lambdatt} \normsq{\gradt{\xt}} \right] + \sum_{t = 1}^{T} \left[\frac{\sigma_t}{2} \normsq{\x_t - \x_t} \right] + \psi_{T+1}(\xopt) \\ & = \sum_{t = 1}^{T} \left[\frac{1}{2 \lambdatt} \normsq{\gradt{\xt}} \right] + \sum_{t = 1}^{T} \frac{\sigma_t}{2} \normsq{\xopt - \xt}. 
\intertext{Since the iterates are bounded on $\cW$, $\norm{\x - \xopt} \leq D$ for all $\x$,}
R(T) & \leq \sum_{t = 1}^{T} \left[\frac{1}{2 \lambdatt} \normsq{\gradt{\xt}} \right] + \frac{D^2}{2} \; \sum_{t = 1}^{T} \sigma_t 
\intertext{Since each $\li$ is $\mu_i$ strongly-convex, $\Ft$ is $\sum_{i = 1}^{t-1} \mu_i$ strongly-convex, and hence $\lambdatt = \sum_{i = 1}^{t} \mu_i$.}
\end{align*}
\end{proof}

\subsection{Proof of Theorem~\ref{thm:ftrl-regret}}
\label{app:ftrl-regret}
\begin{thmbox}
\ftrlregret*
\end{thmbox}
\begin{proof}
In this case, $\mu_i = 0$ i.e. $\li$ is convex only convex without strong-convexity. Using~\cref{lemma:ftrl-main-l2} with $\mu_i = 0$ and defining $\etat := \frac{1}{\sum_{i = 1}^{t} \sigma_t}$,
\begin{align}
R(T) & \leq \sum_{t = 1}^{T} \left[\frac{\etat}{2} \normsq{\gradt{\xt}} \right] + \frac{D^2}{2 \eta_{T}} \label{eq:ftrl-inter}
\end{align}
Recall that we use a constant step-size implying that $\eta_1 = \eta_2 = \eta_T = \eta = \min \left\{\frac{1}{\cE}, \frac{1}{2 L}\right\}$, where $\cE^2 := \sum_{t = 1}^{T} \epsilon_t^2$. With this choice, 
\begin{align*}
R(T)  & \leq \eta L \sum_{t = 1}^{T} \left[\lt(\xt) - \lt(\xopt_t) \right] + \frac{D^2}{2 \eta} \tag{By smoothness, and since $\xopt_t$ is a minimizer of $\lt$.} \\
& = \eta L \sum_{t = 1}^{T} \left[\lt(\xt) - \lt(\xopt) \right] + \eta L \sum_{t = 1}^{T} \left[\lt(\xopt) - \lt(\xopt_t) \right] + \frac{D^2}{2 \eta} \\
& = \eta L \, R(T) + \eta L + \sum_{t = 1}^{T} \epsilon^2_t + \frac{D^2}{2 \eta} \\
\intertext{Since $\eta < \frac{1}{L}$,}
R(T) & \leq \frac{\eta L}{1 - \eta L} \sum_{t = 1}^{T} \epsilon^2_t + \frac{D^2}{2 \eta (1 - \eta L)} 
\intertext{Since $\eta \leq \frac{1}{2L}$, $\frac{1}{1 - \eta L} \leq 2$,}
\implies R(T) & \leq 2 \eta L \sum_{t = 1}^{T} \epsilon^2_t + \frac{D^2}{\eta} \\
\intertext{Since $\eta = \min \left\{\frac{1}{\cE}, \frac{1}{2 L}\right\}$, $\frac{1}{\eta} = \max \left\{\cE, 2L \right\}$,}
R(T) & \leq 2 \eta L \sum_{t = 1}^{T} \epsilon^2_t + D^2 \max \left\{\cE, 2L \right\} 
\leq \frac{2 L}{\cE} \sum_{t = 1}^{T} \epsilon^2_t + D^2 (\cE + 2L) \leq 2 L \cE + D^2 \cE + 2 D^2 L \\
\implies R(T) & \leq 2 D^2 L + (D^2 + 2L) \, \sqrt{\sum_{t = 1}^{T} \epsilon_t^2}
\end{align*}
\end{proof}

\subsection{Proof of Theorem~\ref{thm:adaftrl-regret}}
\label{app:adaftrl-regret}
\begin{thmbox}
\adaftrlregret*
\end{thmbox}
\begin{proof}
We follow the same proof as~\cref{thm:ftrl-regret} until~\cref{eq:ftrl-inter}. Since $\etat = \frac{\alpha}{\sqrt{\sum_{i = 1}^{t} \normsq{\gradi{\x_i}}}}$, 
\begin{align*}
R(T) & \leq \frac{\alpha}{2} \, \sum_{t = 1}^{T} \left[\frac{\normsq{\gradt{\xt}}}{\sqrt{\sum_{i = 1}^{t} \normsq{\gradi{\x_i}}}} \right] + \frac{D^2}{2 \alpha} \sqrt{\sum_{i = 1}^{T} \normsq{\gradi{\x_i}}} \\
\intertext{Bounding $\sum_{t = 1}^{T} \left[\frac{\normsq{\gradt{\xt}}}{\sqrt{\sum_{i = 1}^{t} \normsq{\gradi{\x_i}}}} \right] \leq \sqrt{\sum_{t = 1}^{T} \normsq{\gradi{\x_i}}}$ using the AdaGrad inequality in~\citep{duchi2011adaptive,levy2018online},}
R(T) & \leq \left(\frac{\alpha}{2} + \frac{D^2}{2\alpha} \right) \, \sqrt{\sum_{t = 1}^{T} \normsq{\gradi{\x_i}}} \\
\intertext{By smoothness, and since $\xopt_t$ is the minimizer of $\lt$.}
& \leq  \sqrt{2L} \,\left(\frac{\alpha}{2} + \frac{D^2}{2\alpha} \right) \, \sqrt{\sum_{t = 1}^{T} [\lt(\xt) - \lt(\xopt_t)],} \\   
& = \sqrt{2L} \left(\frac{\alpha}{2} + \frac{D^2}{2\alpha} \right)  \, \sqrt{\sum_{t = 1}^{T} [\lt(\xt) - \lt(\xopt) + \lt(\xopt) -\lt(\xopt_t)]} \\
\intertext{Recall that $\epsilon^2_t := \lt(\xopt) -\lt(\xopt_t)$, and using the definition of $R(T)$,}
\implies \sum_{t = 1}^{T} [\lt(\xt) - \lt(\xopt)] & \leq \sqrt{2L} \left(\frac{\alpha}{2} + \frac{D^2}{2\alpha} \right)  \, \sqrt{\sum_{t = 1}^{T} [\lt(\xt) - \lt(\xopt)] + \sum_{t=1}^{T} \epsilon^2_t} \\
\intertext{Squaring both sides,}
\left(\sum_{t = 1}^{T} [\lt(\xt) - \lt(\xopt)] \right)^2 & \leq 2L \left(\frac{\alpha}{2} + \frac{D^2}{2\alpha} \right)^2  \left(\sum_{t = 1}^{T} [\lt(\xt) - \lt(\xopt)] + \sum_{t=1}^{T} \epsilon^2_t \right) \\
\intertext{Using~\cref{lem:quadratic-inequality},}
R(T) = \sum_{t = 1}^{T} [\lt(\xt) - \lt(\xopt)] & \leq 2L \left(\frac{\alpha}{2} + \frac{D^2}{2\alpha} \right)^2 + \sqrt{2L} \left(\frac{\alpha}{2} + \frac{D^2}{2\alpha} \right) \sqrt{\sum_{t=1}^{T} \epsilon^2_t} 
\end{align*}
\end{proof}

\begin{thmbox}
\begin{lemma}
\label{lem:quadratic-inequality}
If $x^2 \leq a(x+b)$ for $a \geq 0$ and $b \geq 0$,
\begin{align*}
x \leq \frac{1}{2}\sqrt{a^2 + 4ab} + a \leq a + \sqrt{ab}. 
\end{align*}
\end{lemma}
\end{thmbox}
\begin{proof}
The starting point is the quadratic inequality
$x^2 - ax - ab \leq 0$.
Letting $r_1 \leq r_2$ be the roots of the quadratic, 
the inequality holds if $x \in [r_1, r_2]$. 
The upper bound is 
then given by using $\sqrt{a+b} \leq \sqrt{a} +\sqrt{b}$
\begin{align*}
r_2  = \frac{a + \sqrt{a^2 + 4ab}}{2} \leq \frac{a + \sqrt{a^2} + \sqrt{4ab}}{2} = a + \sqrt{ab}. 
\end{align*}
\end{proof}

\subsection{FTRL in the non-smooth, but Lipschitz setting}
\label{app:ftrl-nonsmooth}
In the absence of smoothness, we will make the standard assumption that each $\li$ is $G$-Lipschitz, meaning that for all $\x$, $\norm{\nabla \li(\x)} \leq G$.

\begin{thmbox}
\begin{restatable}[FTRL - Lipschitz + Convex]{theorem}{ftrlregretns}
Assuming each $l_t$ is (i) $G$-Lipschitz, (ii) convex, FTRL with $\etat = \frac{\alpha}{\sqrt{t}}$ achieves the following regret,  
\begin{align*}
R(T) & \leq \frac{\sqrt{T}}{2} \left[G^2 \alpha + \frac{D^2}{\alpha} \right] 
\end{align*}
where $D$ is the diameter of $\cW$. 
\label{thm:ftrl-regret-ns}
\end{restatable}
\end{thmbox}

\begin{proof}
Using~\cref{lemma:ftrl-main-l2} when $\mu_i = 0$, defining $\etat := \frac{1}{\sum_{i = 1}^{t} \sigma_t}$ and bounding $\normsq{\nabla \lt(\xt)} \leq G^2$, 
\begin{align*}
R(T) & \leq \frac{G^2}{2} \sum_{t = 1}^{T} \left[\etat \right] + \frac{D^2}{2 \eta_{T}} \\    
\intertext{For $\etat = \frac{\alpha}{\sqrt{t}}$,}
R(T) & \leq \frac{G^2 \alpha \sqrt{T}}{2} + \frac{D^2 \sqrt{T}}{2 \alpha}
\end{align*}
\end{proof}
\newpage
\begin{thmbox}
\begin{restatable}[AdaFTRL - Lipschitz + Convex]{theorem}{adaftrlregretns}
Assuming each $l_t$ is (i) $G$-Lipschitz, (ii) convex, AdaFTRL with $\etat = \frac{\alpha}{\sqrt{\sum_{i = 1}^{t} \normsq{\gradi{\x_i}}}}$ achieves the following regret,  
\begin{align*}
R(T) & \leq \left(\frac{\alpha}{2} + \frac{D^2}{2\alpha} \right) \, G \sqrt{T}
\end{align*}
where $D$ is the diameter of $\cW$. 
\label{thm:adaftrl-regret-ns}
\end{restatable}
\end{thmbox}
\begin{proof}
We follow the same proof as~\cref{thm:adaftrl-regret}. Since $\etat = \frac{\alpha}{\sqrt{\sum_{i = 1}^{t} \normsq{\gradi{\x_i}}}}$, 
\begin{align*}
R(T) & \leq \left(\frac{\alpha}{2} + \frac{D^2}{2\alpha} \right) \, \sqrt{\sum_{t = 1}^{T} \normsq{\gradi{\x_i}}} 
\end{align*}
Bounding $\normsq{\nabla \lt(\xt)} \leq G^2$ completes the proof. 
\end{proof}

\subsection{FTL in the smooth, strongly-convex setting}
\label{app:ftlregret-sc}
\begin{thmbox}
\begin{restatable}[FTL - Smooth + Convex]{theorem}{ftlregretsc}
Assuming that (i) each $\li$ is $\mu$ strongly-convex for $\mu > 0$, (ii) smooth, the regret for the FTL update in~\cref{eq:ftl-update} can be bounded as:
\begin{align*}
R(T) & \leq \frac{DL}{\mu} (1 + \log(T))    
\end{align*}
\label{thm:ftlregretsc}
\end{restatable}
\end{thmbox}
\begin{proof}
FTL can be considered as a special case of the general FTRL update in~\cref{eq:ftrl-update-gen} with $\psit(\x) = 0$ for all $t$ and $\x$, meaning that $\sigmat = 0$. Using~\cref{lemma:ftrl-main-l2} in this case,
\begin{align*}
R(T) & \leq \sum_{t = 1}^{T} \left[\frac{1}{2 \sum_{i= 1}^{t} [\mu]} \normsq{\gradt{\xt}} \right] \\ 
\intertext{Using smoothness and since $\cW$ has diameter $D$,}
& \leq D L \sum_{t = 1}^{T} \frac{1}{2 \mu t} \leq \frac{DL}{\mu} (1 + \log(T)).
\end{align*}
\end{proof}

\subsection{FTL in the strongly-convex, non-smooth, but Lipschitz setting}
\label{app:ftlregret-scns}
\begin{thmbox}
\begin{restatable}[FTL - Lipschitz + Convex]{theorem}{ftlregretscns}
Assuming that (i) each $\li$ is $\mu$ strongly-convex for $\mu > 0$, (ii) $G$-Lipschitz, the regret for the FTL update in~\cref{eq:ftl-update} can be bounded as:
\begin{align*}
R(T) & \leq \frac{G^2}{2 \mu} (1 + \log(T))  
\end{align*}
\label{thm:ftlregretscns}
\end{restatable}
\end{thmbox}
\begin{proof}
Following the same proof as~\cref{thm:ftlregretsc},
\begin{align*}
R(T) & \leq \sum_{t = 1}^{T} \left[\frac{1}{2 \mu t} \normsq{\gradt{\xt}} \right] \\ 
\intertext{Since $\normsq{\gradt{\xt}} \leq G^2$,}
R(T) & \leq \frac{G^2}{2 \mu} (1 + \log(T)).
\end{align*}

\end{proof}

%% file: app-proofs-ftl.tex
\section{Proof of Theorem~\ref{thm:ftl} }
\label{app:ftl-proofs}

\begin{thmbox}
\ftloil*
\
\end{thmbox}

\begin{proof} 
Recall that at every round FTL returns the following set of parameters:
\begin{equation}
    \pi_T = \argmin_\pi F_{T-1}(\pi) = \argmin_\pi \sum_{t=1}^{T-1} l_t(\pi) = \argmin_\pi \sum_{t=1}^{T-1} \mathbb{E}_{d^\gamma_{\pi_{t}}}[D(\pi, \pi_e)]
\end{equation} 
The loss at round $t$ can be decomposed using the marginal state distribution at round t. Specifically, if $p_{t-1}^\tau(s)$ is the probability of visiting state $s$ at time-stamp $\tau$ under the policy $\pi_{t-1}$, then, 
\begin{align}
    %  l_t(\pi) &= \sum_{m=0}^\infty l_t^m(\pi) = (1-\gamma) \sum_{m=0}^\infty \gamma^m \int_s p_{\pi_{w_{t-1}}}^m(s) D[\pi(\cdot|s),\pi_e(\cdot|s)]ds \\
    %  l_t^m(\pi) & := (1-\gamma) \gamma^m \int_s p_{\pi_{w_{t-1}}}^m(s) D[\pi(\cdot|s),\pi_e(\cdot|s)]ds
    % l_t(\pi) 
    % &= \sum_{m=0}^\infty l_t^m(\pi)
    % = (1-\gamma) \sum_{m=0}^\infty \gamma^m \mathop{\mathbb{E}}_{p^{\tau}_{\pi_{w_{t}}}}
    % [D(\pi(\cdot|s),\pi_e(\cdot|s))] \\
    l_t(\pi_t) &=  (1-\gamma) \sum_{\tau=t}^\infty \gamma^\tau\mathop{\mathbb{E}}_{p^{\tau}_{t}}
    [D(\pi_t(\cdot|s),\pi_e(\cdot|s))] + 
     (1-\gamma) \sum_{\tau=0}^{t-1} \gamma^\tau \mathop{\mathbb{E}}_{p^{\tau}_{t}}
    [D(\pi_t(\cdot|s),\pi_e(\cdot|s))] 
\end{align}
From~\cref{lem:interp_occ}, we get that for all $\tau < t$ we have $p^{\tau}_{t} = p^{\tau}_{\pi_{e}}$. This implies:
\begin{align}
    l_t(\pi_t) &=  (1-\gamma) \sum_{\tau=t}^\infty \gamma^\tau\mathop{\mathbb{E}}_{p^{\tau}_{t}}
    [D(\pi_t(\cdot|s),\pi_e(\cdot|s))] + 
     (1-\gamma) \sum_{\tau=0}^{t-1} \gamma^\tau \mathop{\mathbb{E}}_{p^{\tau}_{\pi_{e}}}
    [D(\pi_t(\cdot|s),\pi_e(\cdot|s))] 
\end{align}
In settings where we are able to exactly match the expert, by using FTL, $\pi_{t}(\cdot|s) = \pi_e(\cdot|s)$ for all states where $p^\tau_{i}(s) > 0$ for any $i = 1, \ldots, t-1$ and $\tau < t$. Using~\cref{lem:interp_occ}, for $\tau < i$, since $p^\tau_{i} = p^\tau_{\pi_{e}}$, we can conclude that $\pi_{t}(\cdot|s) = \pi_e(\cdot|s)$ for all states where $p^\tau_{\pie}(s) > 0$ for $\tau < t$. This means that the second term $\mathop{\mathbb{E}}_{p^\tau_{\pi_{e}}} [D(\pi_t(\cdot|s),\pi_e(\cdot|s))] = 0$. 
\begin{align} 
    &= (1-\gamma) \sum_{\tau=t+1}^\infty \gamma^\tau\mathop{\mathbb{E}}_{p^{\tau}_{t}}
    [D(\pi(\cdot|s),\pi_e(\cdot|s))] \\
    &\leq \max \left\{\mathop{\mathbb{E}}_{p^{\tau}_{t}}
    [D(\pi(\cdot|s),\pi_e(\cdot|s))] \right \} \, (1-\gamma) \sum_{\tau=t+1}^\infty \gamma^\tau \\
    &\leq C \gamma^{t+1} 
\end{align}
where $C := \max \left \{\mathop{\mathbb{E}}_{p^{\tau}_{t}} [D(\pi(\cdot|s),\pi_e(\cdot|s))] \right \}$. We can now sum the left and right hand sides over $T$.
\begin{align}
    \sum_{t=1}^T l_t(\pi_{t}) - \sum_{t=1}^T l_{t}(\pi^*) &\leq C \frac{1-\gamma^{T+1}}{1-\gamma} 
    \leq  \frac{C}{1-\gamma} 
\end{align}
\end{proof}
\newpage
\begin{thmbox}
\begin{lemma}[Follow the Leader: Induced State-Distribution Under Interpolation] \label{lem:interp_occ}
Let the sequence of loss functions observed be defined as the following:
\begin{align}
    l_t(\pi) &= \mathop{\mathbb{E}}_{d^\gamma_{\pi_{t}}(s)}
    [D(\pi(\cdot|s),\pi_e(\cdot|s))], \quad l_t^\tau(\pi) = \int_s p_{t}^\tau(s) D(\pi(\cdot|s), \pi_e(\cdot|s)) ds 
\end{align}
Assume that the policy class contains $\pi_e$. If at every round we minimize the following function with respect to $\pi$:
\begin{equation}
    F_t(\pi) = \sum_{k=0}^{t-1} l_k(\pi) = (1-\gamma) \sum_{k=0}^{t-1} \sum_{\tau=0}^{\infty} \gamma^\tau l_k^\tau(\pi)   
\end{equation}
then $\forall$ $\tau<t$ we have that $p_t^\tau(s)=p^\tau_e(s)$.
\end{lemma}
\end{thmbox}
\begin{proof}
We now show inductively that at round t, $p_t(s) := p_{\pi_t}(s)=p_{\pi_{e}}(s) \ \ \forall \tau < t$ under the assumptions stated above. The proof proceeds by induction, starting with our base case.

\textbf{Base-Case: $\tau=0$}\\
We can immediately note that the initial state distribution is independent of policy, and thus following the definition of our generative model $p_{1}^0(s_0)=p_{e}^0(s_0)$. 

\textbf{Base-Case: $\tau=1$}\\
Next we can consider the results of a single round of the algorithm, and its effect on $p_{1}^1(s)$. By following FTL under interpolation we have:
\begin{align}
    &\pi_{1} = \argmin_\pi L_1(\pi) = \argmin_\pi l_0(\pi) \tag{Definition of $L_1$.} \\
    &l_0(\pi_1) = 0 \implies l_0^k(\pi_1) = 0 \ \ \forall k. \tag{Interpolation assumption.}  
    \\
    \implies & D[\pi_1(\cdot|s),\pi_e(\cdot|s)] =0  \quad \forall s \ \text{s.t.} \ p^0_0(s) := p^0_e(s)  > 0 \tag{Definition of $l_0^1$.}
    \\
    \implies& \pi_{1}(a|s) = \pi_e(a|s) \quad \forall s \ \text{s.t.} \ p^0_0(s) > 0, \tag{ Definition of a divergence and assumption stated below.}
\end{align} 
The final line holds provided the divergence $D$ is strongly convex with respect to the policy $\pi$.
% for all actions such that $\pi_e(a|s) > 0$ we also have that $\pi_1(a|s) >0$.
% We note that the final line only follows provided that the distribution of $\pi_1$ supports $\pi_e$, in the sense that if $\pi_e>0\rightarrow\pi_t>0$, or if the divergence used is strongly convex with respect to the policy. T
% The second option is notable different then the usual assumption given by DAGGER and FTL which requires that we are strongly convex with respect to the parameters. 
Now we have that both $p^0_{1}(s)=p^0_{e}(s)$ by $\tau$=0, and $\pi_{1}(a|s) = \pi_e(a|s)$, we can get direct equality with respect to state at the next time-step:
\begin{align}
  p^1_{e}(s') 
  &= \int_{s}\int_{a} p(s'|s,a)\pi_e(a|s)p^0_e(s) dsda \\
  &= \int_{s}\int_{a} p(s'|s,a)\pi_1(a|s)p^0_0(s) dsda \\
  &= p^1_{1}(s')
\end{align}

\textbf{Inductive Step: } \\
Assume that for some arbitrary round t we have the following: $p^{\tau}_{e}(s) = p^{\tau}_{t-1}(s) \ \ \forall \tau < t $, and we want to prove that at round $t+1$, $p^\tau_{e}(s) = p^\tau_{t}(s)  \ \ \forall \tau < t+1 $, \\

(1) Following the same argument as before $\pi_t(a|s) = \pi_e(a|s) \ \ \forall a, \quad \forall s \ \text{s.t.} \ p^{t-1}_e(s) > 0$:
% \begin{align}
% %  \pi_t &= \argmin_\pi L_t (\pi)  \implies L_t(\pi_t) = \sum_{k=0}^{t-1} l_k(\pi_t) = 0 \tag{Interpolation.} \\
% %     \implies& l_k(\pi_t) = 0 \ \ \forall k\leq t-1.\tag{Definition of divergence.}  \\
% %     \implies& D[\pi_t(\cdot|s_{t-1}),\pi_e(\cdot|s_{t-1})] =0 \ \ \forall s_t \ \text{s.t.} \ p(s_t) > 0 \tag{Definition of  $l_{t-1}(\pi_t)$} \\
% %     \implies
%   & \pi_t(a_{t-1}|s_{t-1}) = \pi_e(a_{t-1}|s_{t-1}) \ \ \forall s_{t-1} \ \text{s.t.} \ p_e(s_{t-1}) > 0 \tag{Definition of $D$ and Interpolation.}
% \end{align}

(2) We can show that $p_{t-1}^{t-1}(s)=p_{t}^{t-1}(s) \ \ \forall s_{t-1} \ \text{s.t.} \ p^{t-1}_{e}(s)>0$ by way of contradiction. Consider the case where the above does not hold, but the policy interpolates all previous functions observed. By definition of $\pi_t$ and $\pi_{t-1}$ for all $\tau<t$ if $p^\tau_e(s)>0$ then $\pi_t(\cdot|s)=\pi_e(\cdot|s)$ and $\pi_{t-1}(\cdot|s)=\pi_e(\cdot|s)$.

% For $p_{t}^{t-1}(s)\neq p_{t-1}^{t-1}(s)$, we would require that require that for some $m\leq t-1$,
If $p_{t-1}^{t-1}(s)\neq p_{t}^{t-1}(s) \ \ \forall s_{t-1} \ \text{s.t.} \ p^{t-1}_{e}(s)>0$, then let us consider the first time step $\tau^*$ such that $p_{t-1}^{\tau^*}(s)\neq p_{t}^{\tau^*}(s) \ \ \forall s \ \text{s.t.} \ p^{\tau^*}_{t-1}(s) = p^{\tau^*}_{e}(s)>0$. Then by definition of $\tau^*$ and the inductive hypothesis, $p_{t-1}^{\tau^*-1}(s)=p_{t}^{\tau^*-1}(s) \ \ \forall s \ \text{s.t.} \ p^{\tau^*-1}_{t-1}(s) = p^{\tau^*-1}_{t}(s) = p^{\tau^*-1}_{e}(s)>0$. Now by the definition of the generative model this implies that  $\exists s \ \ \text{s.t.}\ \ \pi_t(\cdot|s)\neq \pi_{t-1}(\cdot|s) \ \text{and} \ p^{\tau^*-1}_{e}(s)>0$. This is a contradiction, as we showed above. Hence $p_{t-1}^{t-1}(s)=p_{t}^{t-1}(s) \ \ \forall s_{t-1} \ \text{s.t.} \ p^{t-1}_{e}(s)>0$. 

% $p_{{t-1}}(s_{m})!=p_{\pi_{e}}(s_{m})$. Specifically, consider the first instance where this equality is broken and call it $m^*$. At time index $m^*$, we would have that at time index $m^*-1$, one of three things occurred: (i) $p_{\pi_{t-1}}(s_{m^*-1})$ $!=p_{\pi_{e}}(s_{m^*-1})$, which cannot be true by definition of $m^*$ (ii) $p_{\pi_{t-1}}(s_{m^*}|a_{m^*-1}, s_{m^*-1})$, which cannot be possible following the definition of p, (iii) $p_{\pi_{t-1}}(a_{m^*-1}|s_{m^*-1})!=p_{\pi_{e}}(a_{m^*-1}|s_{m^*-1})$ which also cannot occur following our interpolation assumption. Therefore we have reached a contradiction, the index $m^*$ cannot exist given the assumptions placed on the problem, and thus we must have that  $p_{\pi_{t}}(s_{t-1})=p_{\pi_{t-1}}(s_{t-1}) \ \ \forall s_{t-1} \ \text{s.t.} \ p_e(s_{t-1})$. \\
We can again put all this together to get the desired result:
\begin{align}
    p^{t}_{t}(s^\prime) 
    &= \int_{s}\int_{a} p(s'|s,a) \pi_t(a|s) p^{t-1}_{t}(s)  ds da  \tag{Definition of Marginal.}\\
    &= \int_{s}\int_{a} p(s'|s,a) \pi_t(a|s) p^{t-1}_{t-1}(s) ds da \tag{Sub-proof (2).} \\
    &= \int_{s}\int_{a} p(s'|s,a) \pi_t(a|s) p^{t-1}_{e}(s) ds da \tag{Inductive hypothesis.} \\
    &= \int_{s}\int_{a} p(s'|s,a) \pi_e(a|s) p^{t-1}_{e}(s) ds da \tag{FTL} \\
    &= p^t_{e}(s'). \tag{Definition of Marginal.} 
\end{align}
Therefore by induction we have that under interpolation, FTL ensures that at round t $p_{t}^\tau(s)=p_{e}^\tau(s)\forall \tau \leq t-1$. 
\end{proof}

%% file: app-experimental-details.tex
\section{Additional Experimental Details}
\label{app:exp-details}

\subsection{Grid-world Example From Section 1} \label{app:examples:gridworld}
In this experiment, we compare three algorithms: online gradient decent, follow-the-leader, and follow-the-regularized-leader. In this setting we create a $7 \times 7$ grid where the agent is able to move in one of five ways at each time-step: up, down, left, right, or not at all. The expert actions switch in each round with up and right in odd rounds and down and left on even rounds~\cref{fig:adversarial_grid_plots}. The experiment is run for a total of 100 rounds, each of which has a horizon of 5 environment interactions. At the beginning of each round, the agent starts at a position sampled uniformly at random. We verify that FTL incurs substantially larger regret over 1000 rounds. In~\cref{fig:adversarial_grid_plots}, we only include the first 100 rounds to reduce the computational burden. 

To solve the subproblems required for the FTL and the FTRL updates, we use an Armijo, backtracking line-search starting with a fixed step-size~\citep{armijo1966minimization}. For OGD and FTRL, we use a grid-search over the $\eta \in [10^{-5}, \ldots, 10^{5}]$. For a full list of parameters and code, see the \emph{Gridworld-Example folder} inside the accompanying code-base. We also include a combined figure below illustrating the relationship between the reward, which is defined as zero if the agent did not chose the expert action, and one otherwise. 

\begin{figure*}[htbp]
    \centering 
    \includegraphics[width=\textwidth]{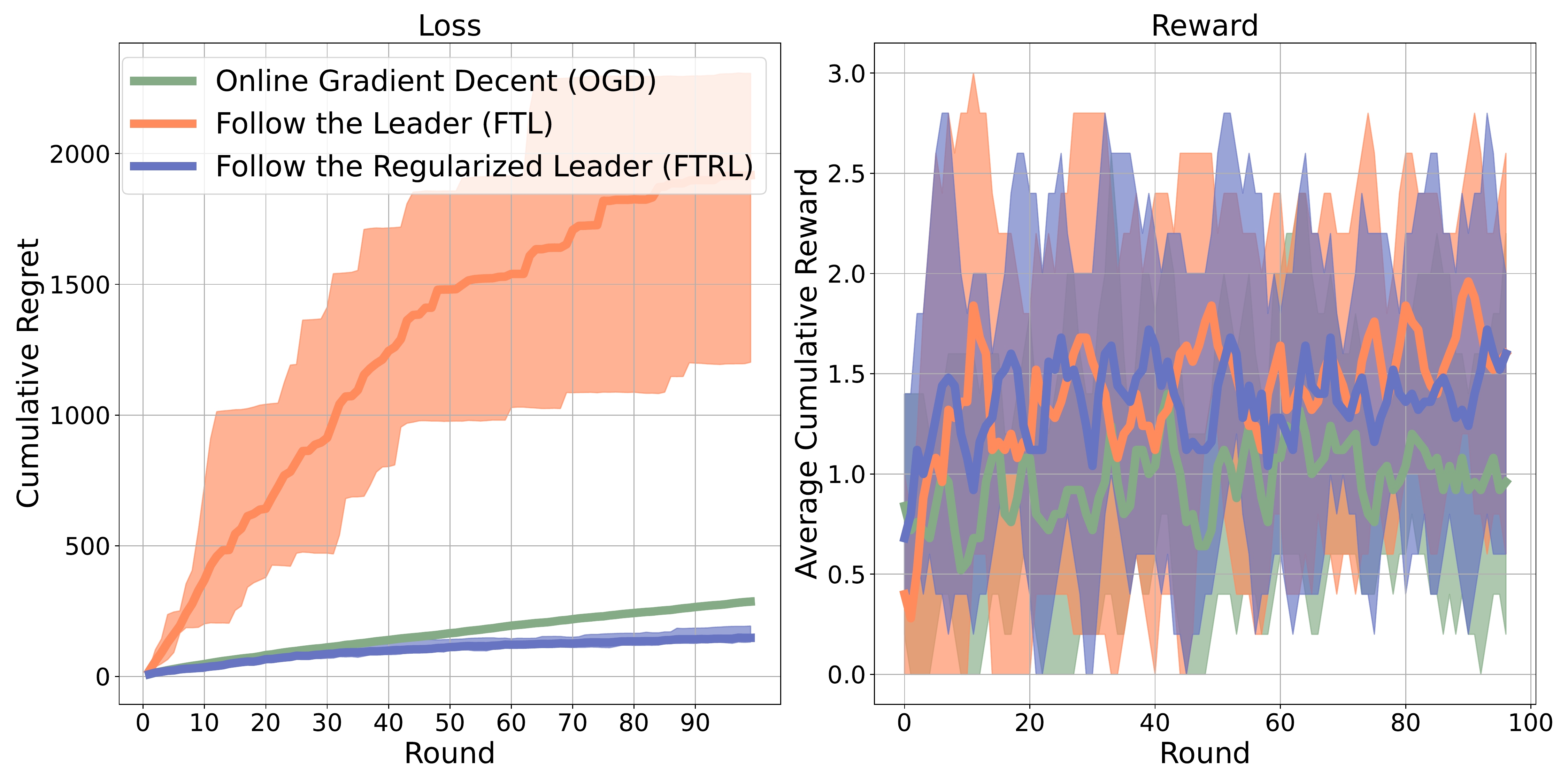}
    \label{fig:app:adversarial_grid_plots}
    \caption{Displays the cumulative regret, and the cumulative reward over 100 rounds.}
\end{figure*}

\subsection{Toy Experiments}
\label{app:toy-experiments}
In these experiments, we consider a series of online regression problems which constitute a mixture of continuous state and discrete action problems (with a logistic loss), and continuous state and continuous action problems (with L1, L2 losses). For setting the hyper-parameters for these problems, we did a grid-search for each problem, loss class, and algorithm, for the initial ten rounds. We then used the best hyper-parameter configuration (in terms of the cumulative regret), and evaluated it for 250 rounds. Similar to~\cref{app:examples:gridworld}, we use a back-tracking line-search. We analyze the performance of each algorithm in two settings for each loss -- \emph{simple} and \emph{adversarial}.

In the simple setting, we construct a random weight matrix $W^* \in \mathbb{R}^{3 \times 10}$ and a random feature matrix $X$ where each state is represented by a $10$-dimensional vector. The action space consists of $3$ actions generated as $W^*X$. For the discrete action case, the logits for each action are proportional to $W^*X$. In the adversarial setting, we use the same generative process as the simple setting, but switch between using $W^{*\top}$ and $-W^{*\top}$ in alternate rounds. For both settings, we sample one environment interaction/round. For a full list of parameters and code, see the \emph{Toy-Experiments folder} inside the accompanying code-base. 

\begin{figure*}[htbp]
    \centering 
    \includegraphics[width=\textwidth]{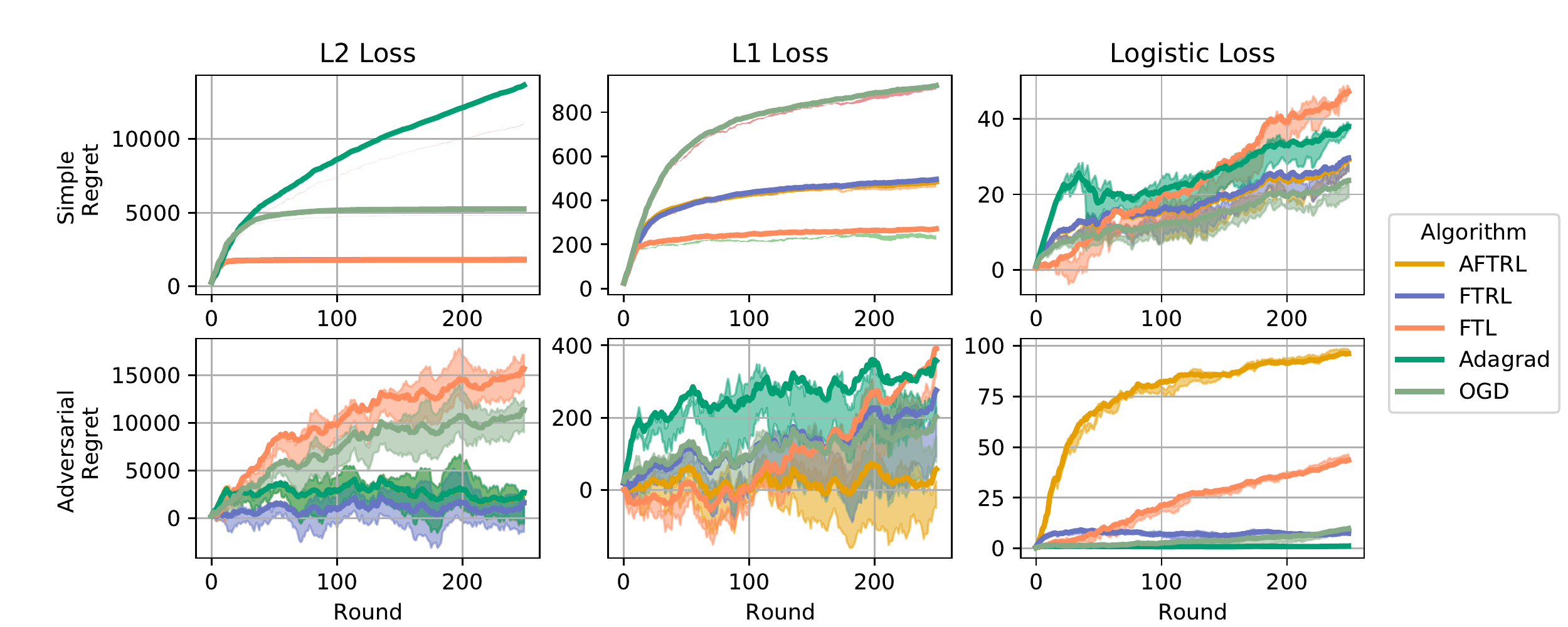}
    \label{fig:toy_experiments}
    \caption{Displays a set of toy experiments in which different online learning algorithms are applied to various online regression problems. In this case we display both the L1 and L2 loss as well as the logistic loss. Our code also includes results for the Huber loss which were omitted. }
\end{figure*}

The figures in~\cref{fig:toy_experiments} display similar trends -- (i) in the simple (non-adversarial) setting, across all losses, FTRL, AFTRL (AdaFTRL), FTL perform significantly better than AdaGrad, OGD (ii) in the adversarial setting, FTL has poor empirical performance across losses.

\subsection{Mujoco}
\label{app:mujoco-details}
In this section, we discuss the details for the  Mujoco~\citep{todorov2012mujoco} continuous control experiments. For a full list of parameters and code, see the \emph{Mujoco-Experiments folder} inside the accompanying code-base. 

\subsubsection{Behavioral Cloning (Interaction under Expert)} 
\begin{figure}[htbp]
    \centering
    \includegraphics[width=\linewidth]{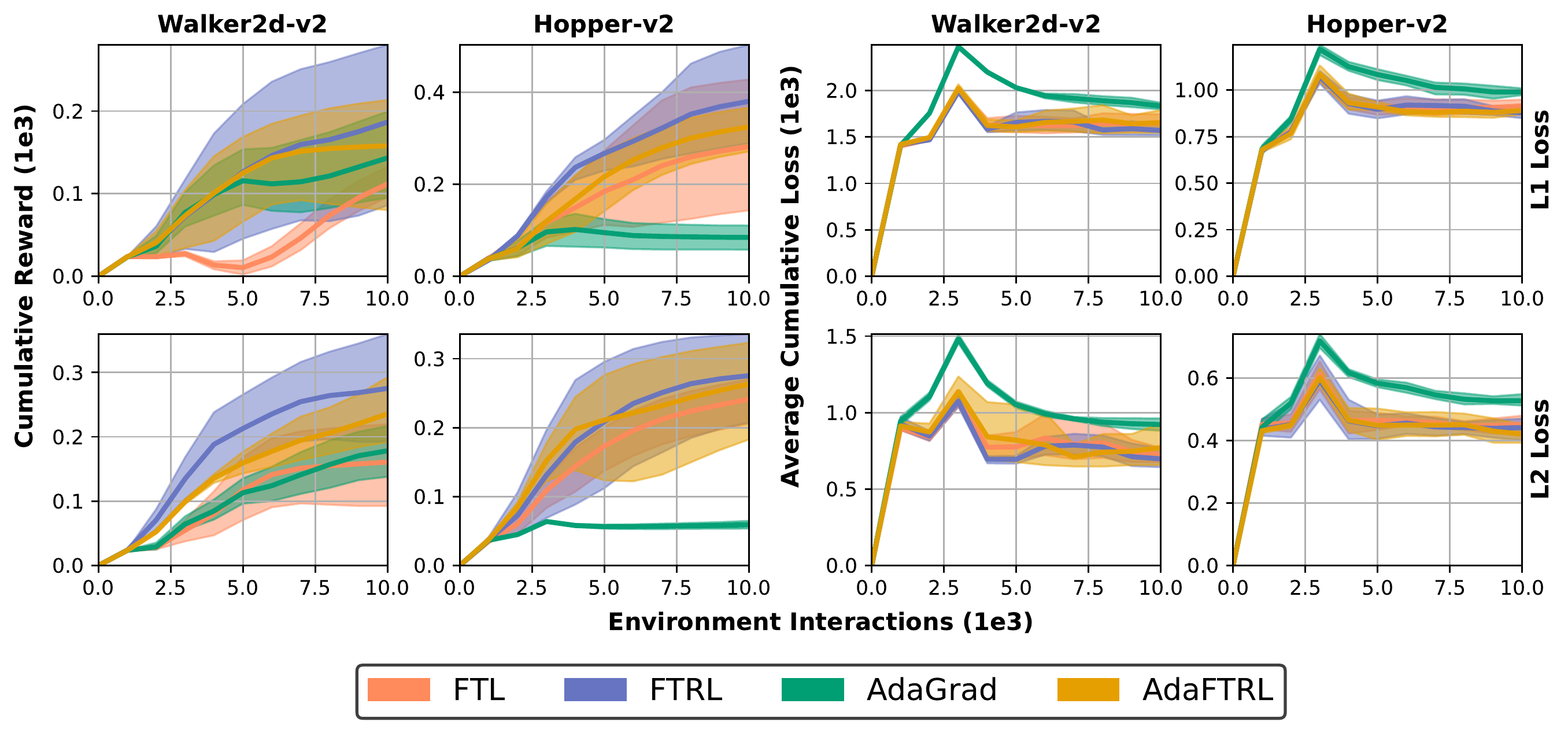}
    \caption{Comparing the algorithms with a linear policy parameterization when using the expert policy to interact with the environment. In this setting, we use 1000 environment interactions per round. We observe that FTRL, AdaFTRL are still the best performing methods, but FTL has slightly worse performance, with AdaGrad performs slightly better.}
    \label{fig:bc_linear}
\end{figure}

\begin{figure}[htbp]
    \centering
    \includegraphics[width=\linewidth]{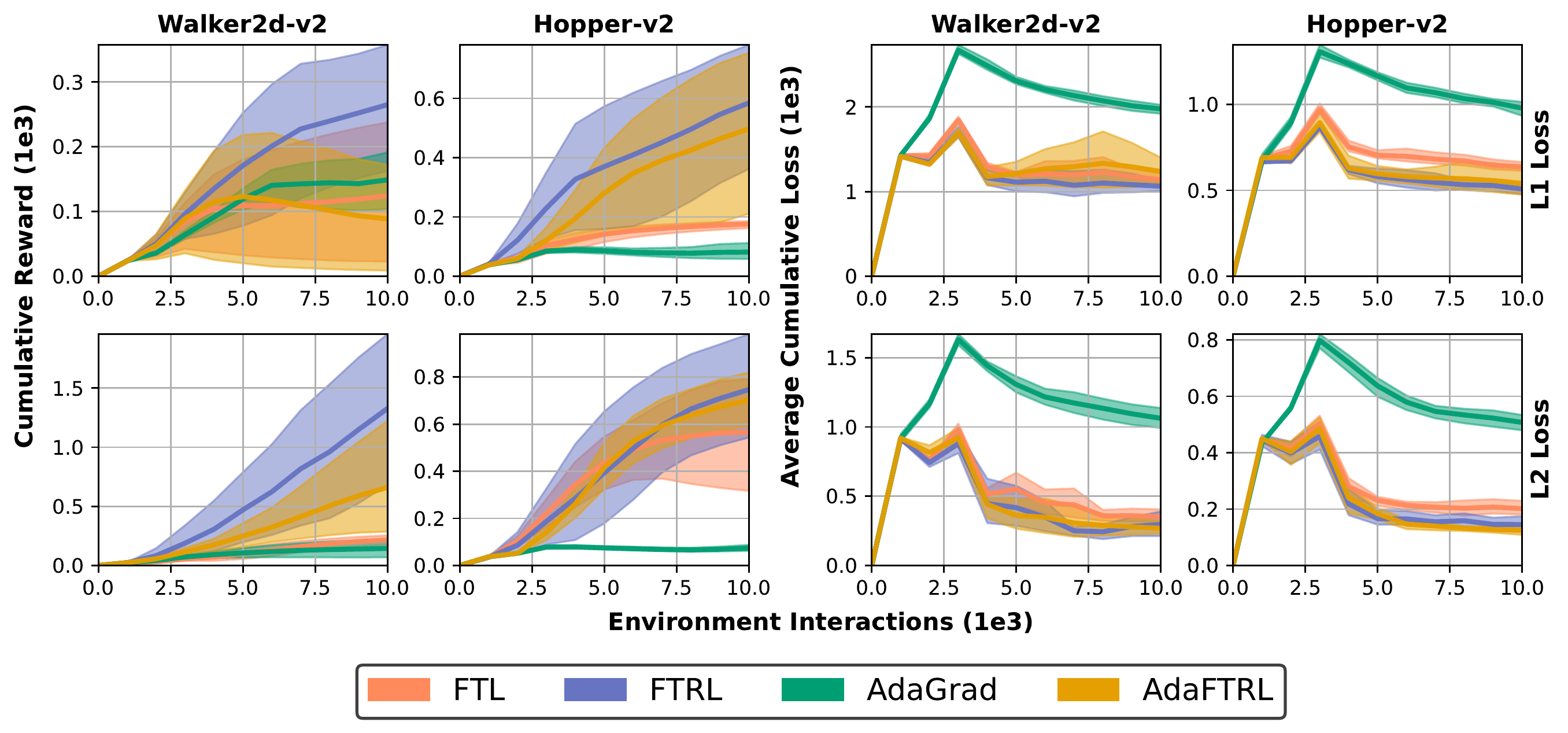}
    \caption{Comparing the algorithms with a neural network policy parameterization when using the expert policy to interact with the environment. In this setting, we use 1000 environment interactions per round. We observe that FTRL, AdaFTRL are still the best performing methods, but both FTL and AdaGrad have significantly worse performance.}
    \label{fig:bc_nn}
\end{figure}

In the experiments presented in the main paper, we interact with the environment using only the current agent policy. Most existing algorithms~\citep{ross2011reduction} use a linear combination of the expert policy and the agents learned policy. We include an ablation in which we only use the expert policy to interact with the environment, and compare the performance of the algorithms for the linear and neural network settings in~\cref{fig:bc_linear} and~\cref{fig:bc_nn} respectively.  


\subsubsection{Decreased sample-sizes for linear models}
\begin{figure}[htbp]
    \centering
    \includegraphics[width=\linewidth]{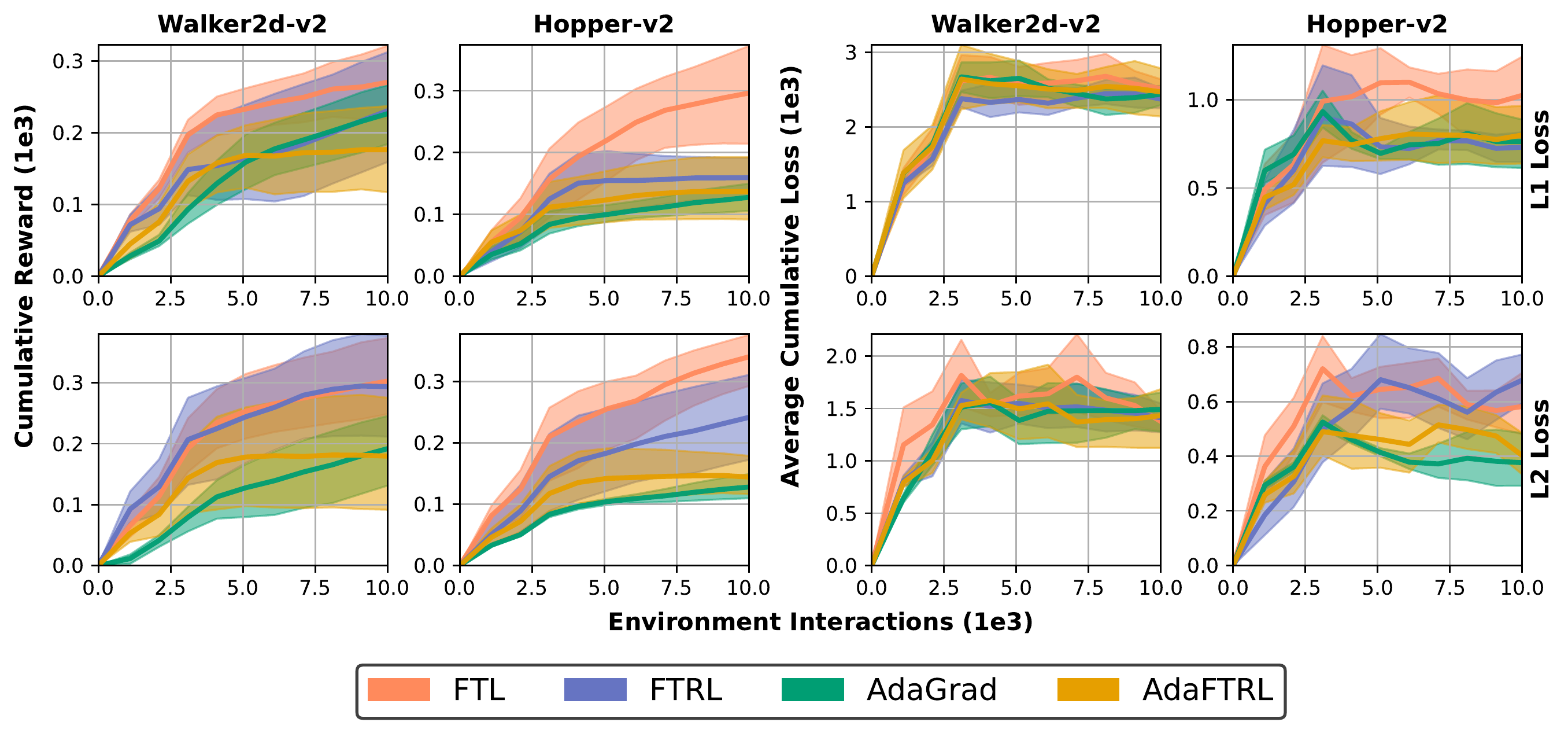}
    \caption{Comparing the algorithms with a linear policy parameterization when using the agent policy to interact with the environment. In this setting, we use 100 environment interactions per round. We observe that FTL, FTRL, AdaFTRL are the best performing methods, while AdaGrad performs better compared to~\cref{fig3:linear}.}
    \label{fig:lowbs_0}
\end{figure}

\begin{figure}[htbp]
    \centering
    \includegraphics[width=\linewidth]{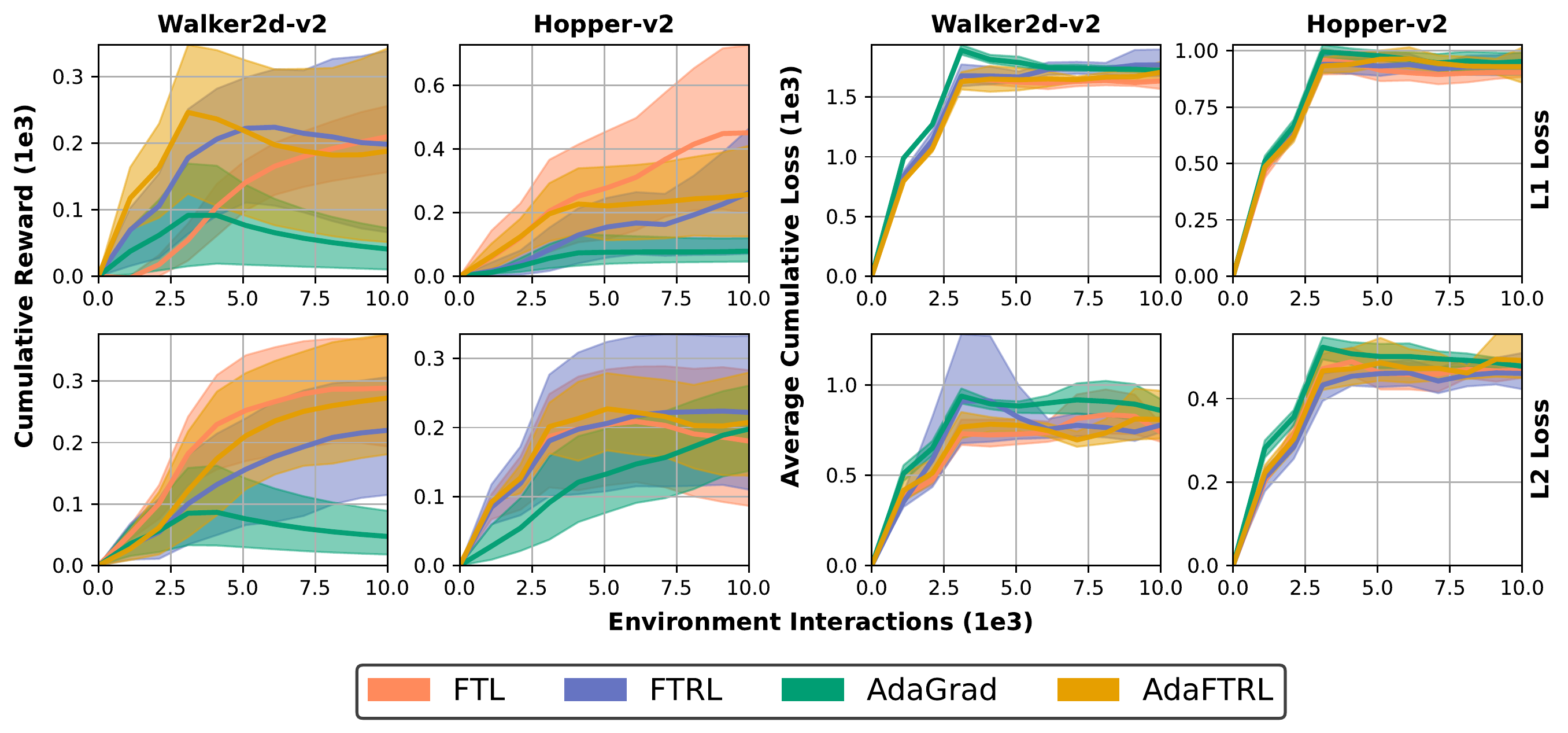}
    \caption{Comparing the algorithms with a linear policy parameterization when using the expert policy to interact with the environment. In this setting, we use 100 environment interactions per round. We observe that FTL, FTRL, AdaFTRL are the best performing methods, while AdaGrad has worse performance.}
    \label{fig:lowbs_1}
\end{figure}

In this section, we demonstrate the effect of using a reduced number of environment interactions per round. We use 100 environment interactions/round and show the results for the linear policy parameterization (\cref{fig:lowbs_0}, \cref{fig:lowbs_1}). 


\subsection{Atari}
\label{app:atari}
As was described in the main paper, in this setting both the expert and learned policy parameterize a categorical distribution which takes an input an 256 by 256 image of the atari screen. Following the same data augmentation as~\citet{schulman2015trust}, we additionally convert the image to a 84 by 84 grey scale image and stack the previous four states observed (also called frame stacking). In this setting the expert is learned using the PPO algorithm, and we again use the architecture described in~\citet{schulman2015trust}. To reduce the computational burden, we only use 3 independent runs. For a full list of parameters and code, see the \emph{Atari-Experiments folder} inside the accompanying code-base. 